\documentclass[11pt]{article}

\usepackage[preprint]{acl}

\usepackage{times}
\usepackage{latexsym}

\usepackage[T1]{fontenc}

\usepackage[utf8]{inputenc}

\usepackage{microtype}

\usepackage{inconsolata}

\usepackage{graphicx}
\usepackage{amsmath}
\usepackage{amssymb}
\usepackage{booktabs}
\usepackage{array}
\usepackage{cleveref}
\AddToHook{cmd/appendix/before}{%
    \crefalias{section}{appendix}%
    \crefalias{subsection}{appendix}
}
\usepackage{subcaption}
\usepackage{tikz}
\usetikzlibrary{arrows.meta,shapes.geometric,positioning,fit,calc,backgrounds}
\usepackage{fontawesome5}
\usepackage[most]{tcolorbox}
\usepackage{tabularx}
\usepackage{multirow}
\usepackage{rotating}
\usepackage{adjustbox}

\usepackage{enumitem}
\usepackage{listings}
\definecolor{tableheader}{HTML}{DDE7F0}
\definecolor{tablerowodd}{HTML}{F7FAFC}
\definecolor{tableroweven}{HTML}{EEF4FB}
\definecolor{tablegrid}{HTML}{B8C8D8}
\definecolor{pipelineNeutral}{HTML}{4D4D4D}
\definecolor{pipelineGreen}{HTML}{59A14F}
\definecolor{pipelineGreenFill}{HTML}{EDF7ED}

\newtcolorbox{practicebox}[1][]{
  colback=tablerowodd,
  colframe=tablegrid,
  fonttitle=\bfseries,
  coltitle=pipelineNeutral,
  title=\faLightbulb\ Practical Implication,
  enhanced,
  attach boxed title to top left={yshift=-3mm, xshift=3mm},
  boxed title style={colback=tableheader, colframe=tablegrid},
  sharp corners,
  boxrule=0.5pt,
  #1
}

\newtcolorbox{takeawaybox}[1][]{
  colback=pipelineGreenFill,
  colframe=pipelineGreen!60,
  fonttitle=\bfseries,
  coltitle=pipelineNeutral,
  title=\faCheckCircle\ Takeaway Message,
  enhanced,
  attach boxed title to top left={yshift=-3mm, xshift=3mm},
  boxed title style={colback=pipelineGreen!20, colframe=pipelineGreen!60},
  sharp corners,
  boxrule=0.5pt,
  #1
}

\usepackage{listings}
\lstdefinestyle{prompttemplate}{
  basicstyle=\ttfamily\footnotesize,
  columns=fullflexible,
  keepspaces=true,
  breaklines=true,
  breakatwhitespace=false,
  showstringspaces=false,
  frame=single,
  framerule=0.3pt,
  rulecolor=\color{tablegrid},
  backgroundcolor=\color{tablerowodd},
  xleftmargin=0pt,
  xrightmargin=0pt,
  aboveskip=0.8\baselineskip,
  belowskip=0.8\baselineskip
}

\title{Model-Based Quality Assessment for Massively Multilingual Parallel Data}

\author{
    Abdelaziz M.A. Ibrahim$^{1,*}$ \quad
    Zihao Li$^{2,*}$ \quad
    J\"org Tiedemann$^2$ \quad
    Shaoxiong Ji$^{3,4}$ \\
    $^1$University of Jyväskylä 
    $^2$University of Helsinki 
    $^3$ELLIS Institute Finland 
    $^4$University of Turku \\
    \texttt{abdelaziz.mabdellatif@gmail.com} \\
    \texttt{\{zihao.li,jorg.tiedemann\}@helsinki.fi} \\
    \texttt{shaoxiong.ji@utu.fi} \\
}

\begin{document}
\maketitle
\def\thefootnote{*}\footnotetext{Equal contribution.}\def\thefootnote{\arabic{footnote}}
\begin{abstract}
Large-scale multilingual bitext often contains two distinct problems: non-parallel sentence pairs and low-quality translations.
We decompose model-based assessment for such data into two independent components: parallelism assessment with multilingual embeddings and reference-free quality estimation (QE).
For parallelism, we benchmark four embedding models on FLORES-200 and BOUQuET retrieval tasks, covering 6,654 source--target directions in our target language-pair inventory.
For QE, we evaluate nine reference-free evaluators on professional FLORES-200 translations across 41,412 ordered source--target directions.
Results show that no model is universally reliable across translation directions.
Naive QE ensembles dilute strong model signals, while documented target-language coverage is strongly associated with higher QE scores.
Overall, these findings suggest that multilingual parallel-data assessment is best approached as a direction-aware routing and calibration problem, where no single universal metric is expected to suffice across all languages.
\end{abstract}

\section{Introduction}

Recent progress in large language models (LLMs) and massively multilingual machine translation has increased the practical reach of language technology, but this progress remains unevenly distributed.
Digital resources and model support are still concentrated in a comparatively small set of high-resource languages, while many of the world's more than 7,000 living languages receive limited technological support \citep{joshi2020state,okoloTano2024_closingGap}.
This digital language divide matters for machine translation (MT) because multilingual systems depend on large amounts of training data, and the languages most in need of improved support are often the ones for which clean parallel data are hardest to obtain.

Large multilingual corpus construction therefore faces a coupled data-availability and data-quality problem.
Web-mined and automatically generated bitexts can expand coverage beyond high-resource languages, but these corpora frequently contain noisy, inconsistent, or low-quality sentence pairs \citep{kreutzer2022quality}.
Some pairs may not be translations of each other at all; others may be broadly equivalent but still contain omissions, additions, mistranslations, or severe fluency problems.
At this scale, manual inspection is infeasible, making it essential to identify automatic model-based signals that remain reliable across thousands of translation directions.

This paper decomposes massively multilingual parallel-data assessment into two independent but complementary components: source--target parallelism and translation quality.
The first component asks whether a source sentence and a target sentence express the same content.
We refer to this as \emph{parallelism assessment} and study it with pretrained multilingual embedding models that assign semantic similarity scores to source--target pairs.
The second component asks whether a candidate translation is fluent and meaning-preserving.
We study this with \emph{reference-free quality estimation} (QE), where an evaluator assigns a score directly to a source sentence and candidate translation without requiring a gold reference at inference time \citep{zhao2024mtqe_survey}.
This property makes reference-free QE suitable for large-scale data selection, where preparing human references for every language pair and domain is not realistic \citep{peter2023no_data_like_better_data}.

This decomposition is useful because parallelism and translation quality are related but distinct properties.
A fluent target sentence can be non-parallel if it expresses different content from the source, while a semantically aligned pair can still contain omissions, additions, mistranslations, or fluency errors.
Consequently, a single generic quality score may obscure different failure modes in multilingual bitext collections.
Embedding-based similarity is naturally suited to semantic alignment, whereas reference-free QE provides a complementary signal for adequacy, fluency, and target-side acceptability.

However, both components become difficult in a massively multilingual setting.
A model that works well for a few high-resource language directions may not remain calibrated across the long tail of multilingual directions.
For parallelism assessment, embedding models may vary substantially in retrieval quality across language pairs, making a single global embedding model unreliable.
For QE, labeled datasets based on Multidimensional Quality Metrics (MQM), Direct Assessment (DA), or post-editing annotations cover far fewer directions than a massively multilingual filtering setting requires \citep{specia2021findings,blain2023findings,fomicheva2022mlqe}.
The results motivate a conceptual shift in how we approach multilingual parallel data assessment, moving toward a direction-aware routing and calibration framework rather than searching for a single universal best model.

This study evaluates both components of the proposed assessment framework.
For parallelism assessment, we benchmark multilingual embedding models as semantic aligners and examine how model selection varies by language direction.
For translation QE, we evaluate reference-free models on FLORES-200 \citep{flores200}, re-purposing professional translations as a high-quality surrogate benchmark over 41,412 ordered translation directions.
This design does not replace MQM, DA, post-editing, or downstream training validation; instead, it tests whether model-based signals behave consistently enough to inform large-scale multilingual corpus assessment.

The research is driven by three main questions:

\begin{itemize}[nolistsep,noitemsep]
  \item \textbf{RQ1} asks how model performance varies by translation direction for the two assessment components: (a) embedding-based parallelism assessment and (b) reference-free QE.
 
 \item \textbf{RQ2} asks whether simple unsupervised ensembles provide a more consistent QE signal across translation directions or can serve as fallback options when single-model routing is unreliable.
 
  \item \textbf{RQ3} asks how QE evaluator behavior changes when the source and target languages are covered or not covered by each model's documented language support.
\end{itemize}

Together, these questions motivate a direction-aware evaluation of both components, where model reliability is assessed by translation direction and interpreted with respect to documented language coverage. This benchmarking provides the empirical basis for a direction-aware routing strategy in massively multilingual assessment.

\section{Problem Setup: Parallelism and Translation Quality}
\label{sec:problem_setup}

\begin{figure}[t] 
    \centering
    \resizebox{0.45\textwidth}{!}{%
      \includegraphics[]{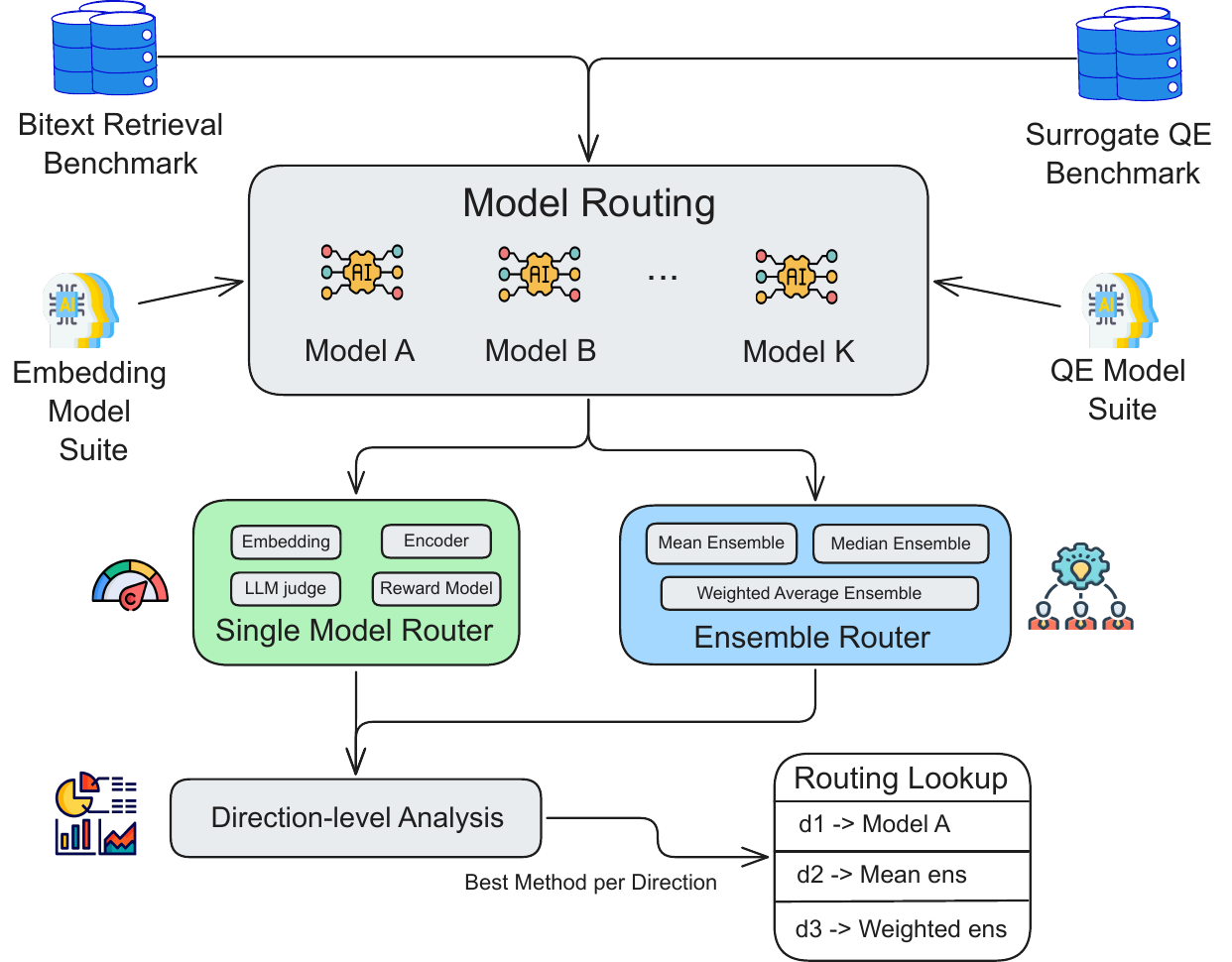}%
    }
    \caption{Unified model-based assessment framework that decomposes quality assessment into two components powered by a single model and an ensemble router. Component 1 assesses source--target parallelism using multilingual embedding models. Component 2 assesses translation quality using reference-free QE models.}
    \label{fig:qe_pipeline}
\end{figure}

We therefore study two independent assessment components: source--target parallelism and reference-free translation quality estimation, illustrated in \Cref{fig:qe_pipeline}.

Let $x$ denote a source sentence in language $\ell_s$ and let $\hat{y}$ denote a candidate target sentence in language $\ell_t$.
A translation direction is denoted as
\[
d = (\ell_s \rightarrow \ell_t).
\]
The central question is whether model-based evaluation can support reliable assessment across a large set of translation directions, particularly for low-resource directions where human-labeled quality data are limited or unavailable.

\subsection{Parallelism and Translation Quality}
\label{sec:parallelism_quality}

We distinguish two related but non-identical properties of a source--target pair.
The first is \emph{parallelism}: whether $x$ and $\hat{y}$ express the same content and can reasonably be treated as translations of each other.
The second is \emph{translation quality}: whether a sentence pair that is likely to be parallel preserves the source meaning fluently and appropriately in the target language.


\subsection{Component 1: Parallelism Assessment}
\label{sec:stage1_setup}

The first component asks whether the source and target sentences express the same meaning.
We model this as semantic similarity in an embedding space.
Given an embedding model $m$, let $e_m(\cdot)$ denote its sentence encoder.
The source and target sentences are encoded as $e_m(x)$ and $e_m(\hat{y})$, and the parallelism score is computed as
\[
a^{(m)}(x,\hat{y}) =
\cos\bigl(e_m(x), e_m(\hat{y})\bigr),
\]
where higher cosine similarity indicates stronger semantic alignment.

For each direction $d$, the parallelism component uses a selected embedding model $m^{\text{align}}_d$ and a direction-specific similarity threshold $\tau_d$.
A sentence pair passes the parallelism component if
\[
\hat{A}(x,\hat{y}) =
\mathbb{I}
\left[
a^{(m^{\text{align}}_d)}(x,\hat{y}) \geq \tau_d
\right].
\]
The threshold $\tau_d$ is direction-specific because embedding similarity distributions may vary substantially across language pairs.

This component is intended as an alignment gate rather than a complete translation-quality metric: it identifies pairs that are likely to be semantically aligned, but it does not by itself determine whether the target sentence is fluent, natural, or error-free.

\subsection{Component 2: Reference-Free Quality Estimation}
\label{sec:stage2_setup}

The second component assesses translation quality for source--target pairs.
We focus on reference-free QE, where an evaluator assigns a score directly to a source sentence and candidate translation without requiring a gold reference at inference time.
A reference-free evaluator $m$ assigns a scalar quality score
\[
q^{(m)}(x,\hat{y}) = f_m(x,\hat{y}),
\]
where higher scores should indicate better translations after normalization.
In this paper, translation quality refers to whether the candidate preserves the source meaning and reads fluently in the target language, including the absence of severe omissions, additions, mistranslations, or local errors \citep{zhao2024mtqe_survey}.

For each direction $d$, the QE component uses a selected evaluator $m^{\text{QE}}_d$ and a direction-specific quality threshold $\gamma_d$.
A sentence pair is retained by the QE component if
\[
\hat{Q}(x,\hat{y}) =
\mathbb{I}
\left[
q^{(m^{\text{QE}}_d)}(x,\hat{y}) \geq \gamma_d
\right].
\]


\section{Direction-Aware Calibration}
\label{sec:direction_aware_calibration}

The two components introduced in \Cref{sec:problem_setup} rely on different model families and scoring functions, but they share the same operational structure.
For both of them, candidate models must be benchmarked by translation direction, a scorer must be selected for each direction, and the resulting scores must be converted into direction-specific thresholds.
This section describes this shared protocol.

Let $z \in \{\text{align}, \text{QE}\}$ denote an assessment component.
For a direction $d=(\ell_s \rightarrow \ell_t)$, let $\mathcal{M}^{z}$ be the set of candidate models for component $z$.
Each model $m \in \mathcal{M}^{z}$ assigns a score to a source--target pair:
\[
s_z^{(m)}(x,\hat{y}) =
\begin{cases}
a^{(m)}(x,\hat{y}), & z=\text{align}, \\
q^{(m)}(x,\hat{y}), & z=\text{QE}.
\end{cases}
\]
Scores from both components are normalized to a common higher-is-better scale, where alignment scores reflect semantic parallelism and QE scores indicate estimated translation quality.
We benchmark models by direction to identify the most reliable signal for each language pair.


\subsection{Parallelism Assessment}
In the parallelism component, for a direction $d=(\ell_s \rightarrow \ell_t)$, the benchmark contains index-aligned source and target sentences.
Given a model $m$, we encode all source sentences and all target sentences, compute the full cosine similarity matrix, and rank target-language candidates for each source sentence.
The correct translation is the target sentence with the same sentence index.
If the correct target sentence for source sentence $i$ has rank $r_i$, the mean reciprocal rank (MRR) is
\[
\mathrm{MRR}_{m,d}
=
\frac{1}{N_d}
\sum_{i=1}^{N_d}
\frac{1}{r_i},
\]
where $N_d$ is the number of benchmark sentence pairs for direction $d$.
Higher MRR indicates that the model more reliably places true translations above non-matching target sentences, and therefore provides a stronger semantic alignment signal for that direction.

We compute this benchmark on multiple parallel evaluation sets.
For model $m$ and direction $d$, the combined parallelism benchmark score is

\[
B^{\mathrm{align}}_{m,d}
=
\frac{1}{|\mathcal{B}^{\mathrm{align}}_{d}|}
\sum_{b \in \mathcal{B}^{\mathrm{align}}_{d}}
\mathrm{MRR}^{(b)}_{m,d}.
\]
This produces a direction-level estimate of how reliably each embedding model retrieves the correct translation across the available benchmark data.

\subsection{Reference-Free Quality Estimation}
For the QE component, reference-free evaluators are compared by their scores on professional FLORES-200 translations rather than by retrieval performance.
For a QE evaluator $m$ and direction $d$, we compute the direction-level mean:
\[
\mu_{m,d}
=
\frac{1}{|I_d|}
\sum_{i \in I_d} q_i^{(m)},
\]
where $I_d$ is the set of examples for direction $d$.
The working assumption is that these human translations are of sufficiently high quality that strong evaluators should assign them scores close to 1.0. 
Since FLORES-200 provides professional translations with uniform per-direction sample sizes, we treat high and stable scores on these translations as a necessary reliability signal.
The overall model score is the macro-average over directions:
\[
\bar{\mu}_m =
\frac{1}{|D|}
\sum_{d \in D} \mu_{m,d},
\]
where $D$ is the set of observed directions.

Thus, both components follow the same direction-level benchmarking principle, but use different signals: retrieval MRR for parallelism, and mean quality scores for QE.

\subsection{Direction-Aware Routing}
\label{sec:routing_concept}

The two components introduced in \Cref{sec:problem_setup} rely on different model families and scoring functions.
The observation that no single model dominates across all directions (as shown in \Cref{sec:results}) motivates a conceptual framework of \emph{direction-aware routing}.
Under this framework, rather than applying a single model to a massively multilingual corpus, one would select the most reliable scorer for each translation direction $d=(\ell_s \rightarrow \ell_t)$ based on empirical benchmark evidence.

For the parallelism component, a routing strategy would select an embedding model $m \in \mathcal{M}^{\mathrm{align}}$ by maximizing the retrieval performance $B^{\mathrm{align}}_{m,d}$ observed on available benchmarks:
\[
m^{\mathrm{align}}_d
=
\arg\max_{m \in \mathcal{M}^{\mathrm{align}}}
B^{\mathrm{align}}_{m,d}.
\]
This approach uses the benchmark results to identify which semantic space is most robust for a given language pair.

Similarly, for the QE component, a routing strategy would select an evaluator $m^{\mathrm{QE}}_d$ based on its performance on professional translations (mean score $\mu_{m,d}$) and diagnostic signals such as documented language coverage.
A simple routing rule would prioritize the evaluator with the strongest direction-level recognition of high-quality translations:
\[
m^{\mathrm{QE}}_d
=
\arg\max_{m \in \mathcal{M}^{\mathrm{QE}}}
\mu_{m,d}.
\]

By framing assessment as a routing problem, we acknowledge that model-based signals are not uniformly calibrated across languages. 
In the following sections, we evaluate the empirical basis for this routing concept by benchmarking how individual models vary across the multilingual inventory.


\section{Component 1: Parallelism Assessment}
\subsection{Embedding Model Suite}
We evaluate four multilingual embedding models as candidate semantic aligners, summarized in \Cref{tab:embedding_models}.

\begin{table}[ht]
\centering
\small
\setlength{\tabcolsep}{4pt}
\begin{tabular}{lll}
\toprule
Short Name & Model ID & Size(M) \\
\midrule
\textsf{Harrier} & \texttt{Harrier-oss-v1-0.6b} & 596 \\
\textsf{mE5-large} & \texttt{multilingual-e5-large} & 560 \\
\textsf{GTE} & \texttt{GTE-multilingual-base} & 305 \\
\textsf{Jina-v3} & \texttt{jina-embeddings-v3} & 570 \\
\bottomrule
\end{tabular}
\caption{
Embedding models used for Component~1:
Harrier \citep{harrier_hf} mE5-large \citep{wang2024multilingual}, GTE \citep{zhang-etal-2024-mgte}, and Jina-v3 \citep{jina-embedding-v3}.
}
\label{tab:embedding_models}
\end{table}





\subsection{Bitext Retrieval Benchmark}
We benchmark the embedding models on two multilingual bitext retrieval datasets: FLORES-200 and BOUQuET$_{\text{Sentence}}$.
FLORES-200 is a sentence-level professionally translated many-to-many MT benchmark covering 204 language varieties, which yields more than 40K ordered translation directions \citep{flores200}.
BOUQuET is a multi-way translation benchmark designed to complement FLORES-style evaluation with broader domain and register coverage.
At the time of experiments, BOUQuET includes 275 completed multi-way parallel languages and provides both sentence- and paragraph-level alignments \citep{andrews-etal-2025-bouquet}.
We use the sentence-level version of BOUQuET because the corpus pairs targeted by our parallelism assessment are sentence-level pairs. 


\section{Component 2: Reference-Free Quality Estimation}

\subsection{QE Model Suite}
We evaluate nine reference-free QE systems, summarized in \Cref{tab:qe_models}.

\begin{table}[ht]
\centering
\small
\setlength{\tabcolsep}{3pt}
\begin{tabularx}{\columnwidth}{@{}l X l@{}}
\toprule
Short Name & Model ID & Category \\
\midrule
\textsf{COMETKiwi} 
& \texttt{COMETKiwi-23-XL} 
& \multirow{2}{*}{Encoder QE} \\

\textsf{xCOMET} 
& \texttt{xCOMET-XL} 
& \\

\midrule
\textsf{MetricX}
& \texttt{MetricX-24-Hybrid-QE} 
& Learned metric \\

\textsf{ReMedy} 
& \texttt{ReMedy-9B} 
& Reward model \\

\midrule
\textsf{M-Prometheus} 
& \texttt{M-Prometheus-7B} 
& \multirow{4}{*}{LLM judge} \\

\textsf{Qwen3-4B}
& \texttt{Qwen3-4B-Instruct-2507} 
& \\

\textsf{Qwen3-8B} 
& \texttt{Qwen3-8B} 
& \\

\textsf{Qwen3-14B} 
& \texttt{Qwen3-14B} 
& \\

\midrule
\textsf{Bicleaner} 
& \texttt{Bicleaner-AI} 
& Cleaner \\
\bottomrule
\end{tabularx}
\caption{
Reference-free QE models used for Component~2:
COMETKiwi \citep{rei2023scaling_up_cometkiwi_wmt}, xCOMET \citep{guerreiro2024xcomet}, MetricX \citep{juraska-etal-2024-metricx}, ReMedy \citep{tan-monz-2025-remedy}, M-Prometheus \citep{pombal2025mprometheussuiteopenmultilingual}, Qwen3 Family \citep{qwen3}, and Bicleaner \citep{zaragoza2022bicleaner}.
}
\label{tab:qe_models}
\end{table}

All LLM evaluators use a shared TASER-style prompt \citep{maheswaran2025taser}, scoring seven quality dimensions and an overall rating on a 0--100 scale (\Cref{app:llm-prompt}).

\subsection{FLORES-200 as a Surrogate QE Benchmark}

FLORES-200 is an MT benchmark rather than a QE-labeled dataset, and it is used here as a high-quality surrogate benchmark for massively multilingual QE comparison \citep{flores200}.
Both the \texttt{dev} and \texttt{devtest} splits are used, producing 83,196,648 source--translation instances after expansion across ordered directions.

The benchmark is interpreted comparatively.
Since FLORES-200 does not provide QE labels, the experiment does not measure correlation with human judgments.
Instead, it compares how strongly each evaluator recognizes professional FLORES translations as high quality across the full multilingual inventory.
This is a narrower claim than full QE validation, but it is directly relevant to filtering: if a model assigns low or unstable scores to professional translations in a direction, its use as an automatic filter for noisier data in that direction becomes difficult to justify.

\subsection{Normalization and Ensembles}

All model outputs are mapped to a common $[0,1]$ range, where higher values indicate better translation quality. 
We normalize \textsf{MetricX} from its 0--25 lower-is-better scale as
$\mathrm{metricx}_{\mathrm{norm}} = 1 - \frac{\mathrm{metricx}}{25}$,
and LLM 0-100 scores as
$\mathrm{llm}_{\mathrm{norm}} = \frac{\mathrm{llm}_{0\text{--}100}}{100}$.
\textsf{Bicleaner}, \textsf{COMETKiwi}, and \textsf{xCOMET} already produce higher-is-better scores on the $[0,1]$ scale.

For RQ2, the benchmark also evaluates unsupervised ensembles as diagnostic baselines for cross-direction consistency and fallback behavior.
These aggregations are not supervised meta-evaluators, and mean or median aggregation should not be expected to outscore the strongest constituent model on a single translation direction.
The question is whether aggregation produces a more stable signal across many directions.
The unrestricted mean, median, and weighted-average ensembles aggregate all available evaluator scores without supervised training.
The weighted ensemble uses the single-model macro-averages as fixed weights, where
$q_i^{(\mathrm{wavg})}=\sum_{m=1}^{M} w_m q_i^{(m)}$ and
$w_m=\bar{\mu}_m/\sum_{j=1}^{M}\bar{\mu}_j$.
Here, $M$ is the number of constituent models and $w_m$ is the normalized weight assigned to model~$m$.
Coverage-aware variants restrict the constituent pool according to documented support for the source language, target language, both languages, or neither language.
Languages are mapped to FLORES-200 codes through exact matching, manually curated aliases, left-trim matching, and qualifier stripping.

\section{Results}
\label{sec:results}

\subsection{RQ1: Single-Model Benchmarking}
\label{sec:benchmark-single-model}

\subsubsection{Parallelism Assessment}
\label{sec:rq1_parallelism}
The two retrieval benchmarks provide parallelism evidence for 6,654 source--target language directions in the corpus we aim to assess.
For each covered direction, we average the FLORES-200 and BOUQuET$_{\text{Sentence}}$ MRR scores when both are available. 
This benchmark allows us to identify, for each direction, the embedding model that would be selected under a routing strategy.

\begin{table}[ht]
\centering
\small
\begin{tabular}{lrr}
\toprule
Model & Avg. MRR $\uparrow$ & Best directions $\uparrow$ \\
\midrule
\textsf{Harrier} & 0.963 & 3,047 \\
\textsf{mE5-large} & 0.953 & 2,013 \\
\textsf{GTE} & 0.903 & 54 \\
\textsf{Jina-v3} & 0.828 & 1,540 \\
\bottomrule
\end{tabular}
\caption{
Parallelism benchmark results for the embedding model suite.
}
\label{tab:parallelism_routing}
\end{table}

Table~\ref{tab:parallelism_routing} shows that \textsf{Harrier} obtains the highest average MRR and is selected for the largest number of directions.
\textsf{mE5-large} is close in average MRR and is routed to 2,013 directions, indicating that it remains highly competitive across the multilingual inventory.
\textsf{Jina-v3} has a lower average MRR overall, but it is still selected for 1,540 directions, showing that it provides the strongest alignment signal for a substantial subset of language pairs.
\textsf{GTE} is selected for only 54 directions, suggesting that it is rarely the top model under the direction-aware routing criterion.

These results show that parallelism assessment would likely benefit from direction-aware model selection.
Although \textsf{Harrier} is the strongest global choice, no single embedding model dominates all covered directions.
This empirical variance provides a strong justification for a direction-aware routing strategy rather than applying one model uniformly to an entire multilingual corpus.

\subsubsection{Reference-Free Quality Estimation}
\label{sec:rq1_qe}

Table~\ref{tab:single_model_results} reports the main single-model comparison over all 41,412 ordered FLORES-200 directions.
The table distinguishes three criteria: first-place frequency, aggregate score, and rank stability.
\textsf{ReMedy} wins the largest number of directions, with 16,367 wins (39.52\%).
\textsf{MetricX} has the highest macro-average, 0.6228.
\textsf{Qwen3-4B} has the best rank profile, with the lowest rank mean (2.39) and rank standard deviation (1.25).

\begin{table}[ht]
  \centering
  \small
  \setlength{\tabcolsep}{3pt}
  \begin{tabularx}{\columnwidth}{@{}Xccc@{}}
    \toprule
    Model & Wins $\uparrow$ & Macro $\uparrow$ & Rank $\downarrow$ \\
    \midrule
    \textsf{ReMedy} & 16.4k (39.5) & 0.5489 & $3.72{\pm}2.89$ \\
    \textsf{Qwen3-4B} & 12.0k (29.0) & 0.6160 & $2.39{\pm}1.25$ \\
    \textsf{MetricX} & 8.8k (21.3) & 0.6228 & $3.16{\pm}1.79$ \\
    \textsf{M-Prometheus} & 1.8k (4.4) & 0.4751 & $5.41{\pm}2.24$ \\
    \textsf{Qwen3-8B} & 1.7k (4.1) & 0.5517 & $3.79{\pm}1.49$ \\
    \textsf{Qwen3-14B} & 0.5k (1.1) & 0.4879 & $5.41{\pm}1.87$ \\
    \textsf{Bicleaner} & 0.2k (0.5) & 0.2141 & $8.18{\pm}1.44$ \\
    \textsf{COMETKiwi} & $<$0.1k (0.1) & 0.3284 & $6.49{\pm}1.74$ \\
    \textsf{xCOMET} & $<$0.1k (0.1) & 0.3221 & $6.45{\pm}1.73$ \\
    \bottomrule
  \end{tabularx}
  \caption{
  QE benchmark results for single-model.
  Wins report first-place count and percentage.
  Rank reports mean $\pm$ standard deviation.
  }
  \label{tab:single_model_results}
\end{table}


The strongest three models account for 37,175 direction-level wins, or 89.8\% of the benchmark, but this concentration should not be interpreted as universal dominance.
\textsf{ReMedy} has the most wins but only the fourth-highest macro-average, indicating strong direction-specific peaks.
\textsf{MetricX} wins fewer directions than \textsf{ReMedy} but has the best macro-average, which points to broader aggregate strength.
The low rank variance of \textsf{Qwen3-4B} makes it the most stable near-top single model.

The margin distribution presented in table~\ref{tab:winning_margin_distribution} further shows that many direction-level wins are narrow.
In 20,082 directions (48.49\%), the gap between the best and second-best model is below 0.05.
Only 10,558 directions (25.50\%) have a winning margin of at least 0.10.
Thus, nearly half of all directions are decided by small score differences, which limits the confidence with which any single evaluator can be declared clearly superior.


\begin{table}[ht]
  \centering
  \small
  \begin{tabular}{lrrr}
    \toprule
    Margin summary & $\geq 0.10$ & 0.05--0.10 & $< 0.05$ \\
    \midrule
    Number of directions & 10{,}558 & 10{,}772 & 20{,}082 \\
    Share of directions (\%) & 25.50 & 26.01 & 48.49 \\
    \bottomrule
  \end{tabular}
  \caption{Distribution of direction-level winning margins, defined as the difference between the top-ranked and second-ranked mean scores for each direction.}
  \label{tab:winning_margin_distribution}
\end{table}

\begin{takeawaybox}
  No single model is uniformly best across the two assessment components.
  Parallelism assessment benefits from per-direction embedding selection, while reference-free QE shows different winners depending on whether we prioritize direction-level wins, macro-average score, or rank stability.
\end{takeawaybox}


\Cref{sec:family-analysis} provides a language family analysis of the QE single-model results.

\subsection{RQ2: Unsupervised Ensembles}
\label{sec:benchmark-ensembles}

RQ2 asks whether simple unsupervised aggregation can provide a more reliable QE signal than a direction-aware single-model strategy.
We compare three unrestricted ensembles over all evaluators: mean, median, and macro-weighted average.
We also consider coverage-aware variants that restrict the ensemble pool to evaluators whose documented language coverage includes both the source and target languages (\emph{both-seen ensembles}).

\Cref{tab:ensemble_benchmark} reports the main ensemble results.
The unrestricted mean, median, and weighted-average ensembles reach macro-averages of 0.4630, 0.4842, and 0.5026, respectively.
All three are substantially below the strongest single-model baselines from RQ1, including \textsf{MetricX} (0.6228) and \textsf{Qwen3-4B} (0.6160).
This should not be interpreted as a per-direction failure to beat the best constituent model, since mean and median aggregation are not designed to exceed the strongest scorer on each individual direction.
Rather, the result shows that averaging across the full evaluator pool dilutes the signal from stronger evaluators because weaker models contribute low scores.

\begin{table}[ht]
  \centering
  \small
  \setlength{\tabcolsep}{3pt}
  \begin{tabularx}{\columnwidth}{@{}Xrrr@{}}
    \toprule
    Method & Macro $\uparrow$ & Rank $\downarrow$ & Dir. \\
    \midrule
    \textsf{MetricX} & 0.6228 & 4.17 & 41{,}412 \\
    \textsf{Qwen3-4B} & 0.6160 & 2.77 & 41{,}412 \\
    \textsf{ReMedy} & 0.5489 & 5.69 & 41{,}412 \\
    \midrule
    \textsf{Mean ens.} & 0.4630 & 9.08 & 41{,}412 \\
    \textsf{Median ens.} & 0.4842 & 8.09 & 41{,}412 \\
    \textsf{Weighted ens.} & 0.5026 & 7.03 & 41{,}412 \\
    \midrule
    \textsf{Both-seen mean ens.} & 0.6901 & 8.20 & 19{,}744 \\
    \textsf{Both-seen median ens.} & 0.7135 & 6.40 & 19{,}744 \\
    \textsf{Both-seen weighted ens.} & 0.7179 & 6.51 & 19{,}744 \\
    \textsf{Both-seen Qwen3-4B} & 0.8498 & 2.45 & 14{,}762 \\
    \bottomrule
  \end{tabularx}
  \caption{
  Main unsupervised ensemble results for RQ2.
  ``Dir.'' denotes eligible directions.
  Both-seen ensembles are evaluated only on directions where both languages are documented as covered.
  }
  \label{tab:ensemble_benchmark}
\end{table}

The rank-stability results show that stability alone is not sufficient.
The unrestricted median ensemble has a low rank standard deviation of 1.71, but its mean rank is 8.09 in the expanded method pool.
It is therefore stable mainly because it remains in the middle of the ranking, not because it is consistently near the top.
Coverage-aware both-seen ensembles obtain higher raw macro-averages, with mean, median, and weighted variants reaching 0.6901, 0.7135, and 0.7179, respectively.
However, these values are computed only on coverage-favorable subsets.
On the same both-seen subset, \textsf{Qwen3-4B} reaches 0.8498, remaining clearly ahead of the best both-seen ensemble.
Thus, naive aggregation does not deliver a stronger cross-direction assessment signal than a direction-aware single-model strategy.
Full results for all individual evaluators and ensemble variants are provided in \Cref{tab:ensemble_benchmark_full}.

\begin{takeawaybox}
  Naive unsupervised ensembles dilute top-tier QE signals.
  Direction-aware routing to the best single evaluator is more effective than simple mean, median, or weighted aggregation.
\end{takeawaybox}

\subsection{RQ3: Coverage and Target-Side Asymmetry}
\label{sec:benchmark-coverage}

\begin{figure*}[!ht]
  \centering
  \includegraphics[width=0.84\textwidth]{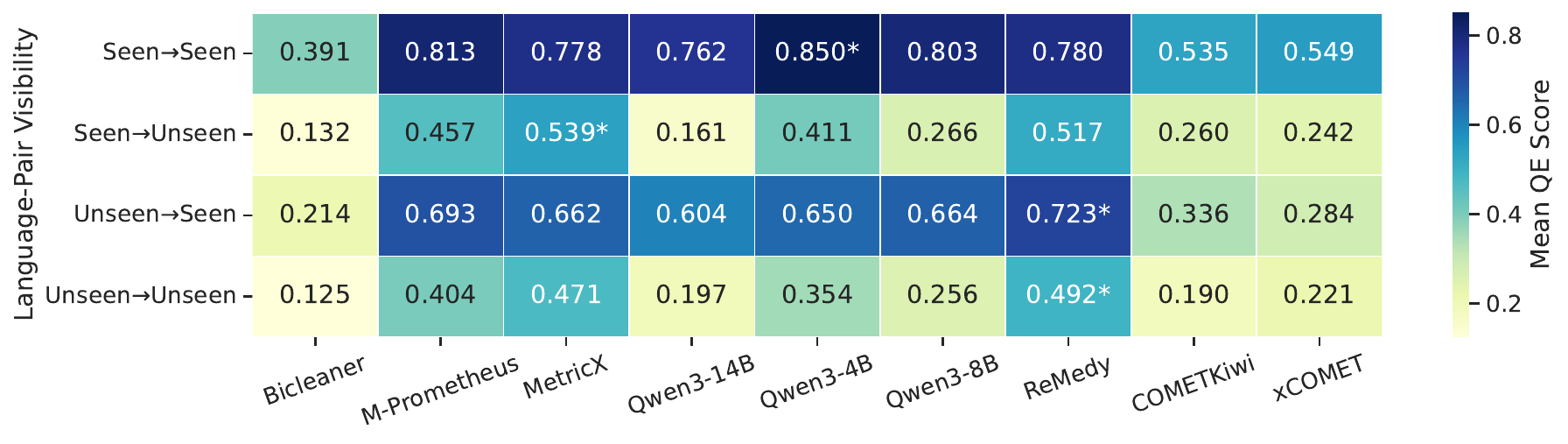}
  \caption{
  Mean QE score by model and source--target coverage condition. 
  Cell values are normalized mean scores; higher values indicate stronger recognition of FLORES translations as high quality.
  Asterisks mark the highest-scoring model within each coverage condition.
  }
  \label{fig:coverage_heatmap}
\end{figure*}

RQ3 examines whether documented language coverage explains part of the direction-level behavior of reference-free QE evaluators.
We group each source--target direction into four visibility conditions according to whether the source and target languages are documented as supported by the evaluator: both seen, source-only seen, target-only seen, or both unseen.
Coverage is only treated as a diagnostic proxy for evaluator reliability, not as proof of training exposure.

Figure~\ref{fig:coverage_heatmap} shows a clear coverage effect.
For every evaluator, the highest mean scores occur when both the source and target languages are documented as supported.
This pattern supports the use of coverage metadata as a routing and confidence signal: directions covered by an evaluator are more likely to receive high scores on professional translations.

The more informative pattern is the asymmetry between source-side and target-side coverage.
In the mixed-coverage conditions, the source-only and target-only subsets contain the same number of directions and are therefore directly comparable.
Across all models, target-only coverage yields higher mean scores than source-only coverage.
For example, \textsf{Qwen3-4B} rises from 0.411 under source-only coverage to 0.650 under target-only coverage, and \textsf{ReMedy} rises from 0.517 to 0.723.
This suggests that reference-free QE is especially sensitive to target-language competence, plausibly because the evaluator must judge the fluency, acceptability, and adequacy of the translated sentence itself.

\begin{takeawaybox}
  Documented language coverage is strongly associated with higher QE scores, and target-language coverage is consistently more important than source-language coverage in mixed-visibility directions.
\end{takeawaybox}

Coverage, however, does not fully solve the routing problem.
Even after selecting the strongest available evaluator for each direction, 7,562 directions (18.3\%) have a best-available mean score below 0.5, and another 3,520 directions (8.5\%) fall between 0.5 and 0.6.
These low-to-moderate scores indicate directions where even professional FLORES translations receive only moderate evaluator scores.
For such directions, automatic QE filtering should be applied conservatively.

\section{Related Work}

Bitext mining aims to identify source--target sentence pairs that are mutual translations in noisy or comparable corpora.
A major line of work uses multilingual sentence embeddings for this purpose: sentences are mapped into a shared space, and embedding similarity is used to filter noisy bitext or retrieve new parallel pairs \citep{schwenk-2018-filtering,artetxe-schwenk-2019-massively}
Later work showed that absolute cosine thresholds can be poorly calibrated across sentences and language pairs, motivating margin-based scoring for more robust parallel sentence retrieval \citep{artetxe-schwenk-2019-margin}.
This embedding-based paradigm has supported large-scale mined corpora such as WikiMatrix and has been extended to unsupervised, contextual, and low-resource settings \citep{schwenk-etal-2021-wikimatrix,keung-etal-2020-unsupervised,heffernan-etal-2022-bitext}.
Our work does not propose a new mining algorithm; instead, we use bitext retrieval benchmarks to select, for each direction, the embedding model that provides the most reliable parallelism signal.

Quality estimation predicts the quality of a source--translation pair without requiring human references at inference time \citep{chatterjee-etal-2018-combining,zhao2024mtqe_survey}. This sentence-level scalar signal is vital for filtering noisy web-mined and synthetic multilingual corpora where reference data is unavailable at scale \citep{peter2023no_data_like_better_data,chaplynskyi2025_parallel_corpus_evaluation_framework}. While recent WMT shared tasks yield strong encoder- and LLM-based evaluators, their validation remains concentrated on a limited set of language directions \citep{blain2023findings,zerva2024findings,lavie2025findings}. Consequently, rather than assuming a single universal evaluator, we treat reference-free QE for massive corpus construction as an empirical model-selection problem. We use human-curated translations as scalable quality anchors to compare evaluators across many directions, rather than as replacements for MQM or Direct Assessment labels.

\section{Conclusion}

We studied model-based quality assessment for massively multilingual parallel data by decomposing it into two independent components: source--target parallelism assessment and reference-free translation quality estimation.

Across both components, the results support a direction-aware view of multilingual data assessment.
Neither embedding-based parallelism assessment nor reference-free QE can be reduced to a single globally optimal model.
Instead, model behavior varies substantially across translation directions, and different evaluation criteria emphasize different aspects of reliability, such as peak performance, average strength, and rank stability.
This suggests that practical multilingual filtering should prioritize direction-level routing and calibration over leaderboard-style model selection.

We also find that simple unsupervised ensembles do not solve the cross-direction reliability problem.
Mean, median, and weighted aggregation dilute strong evaluator signals and do not outperform a direction-aware single-model strategy.
Finally, the coverage analysis shows that documented language support is strongly associated with higher QE scores, with target-language coverage being consistently more important than source-language coverage in mixed-visibility directions.
This suggests that reference-free QE is especially sensitive to target-side competence.

Overall, the results frame massively multilingual parallel-data assessment as direction-aware routing and calibration, with model choice and score interpretation conditioned on the language pair.

\section*{Limitations}

This work does not provide downstream training validation.
We do not claim that applying the proposed routing and filtering strategy necessarily improves MT or LLM training outcomes.
The contribution is instead a benchmark-driven assessment framework for identifying model-based signals that may support large-scale multilingual corpus filtering.
Future work should evaluate whether data selected by these signals improves downstream translation quality, multilingual transfer, or language-model pretraining efficiency.
We also do not evaluate the effect of applying the two components as a cascaded filtering pipeline.
The parallelism and QE components are benchmarked separately, so the results should not be interpreted as evidence that a specific sequential filtering strategy improves corpus quality.

The QE benchmark is a positive-only surrogate evaluation.
FLORES-200 provides professional translations, but it does not include MQM, Direct Assessment, post-editing, or other human QE labels for the full multilingual inventory.
As a result, our QE experiments test whether evaluators assign high and stable scores to high-quality translations, not whether they reliably distinguish all types of noisy, domain-shifted, hallucinated, or partially mistranslated sentence pairs.
Low scores on FLORES are informative as a sign of possible evaluator miscalibration, but high scores do not guarantee that the same evaluator will reject poor translations in web-mined corpora.

The parallelism benchmark is also limited by benchmark coverage and sentence-level assumptions.
FLORES-200 and BOUQuET provide high-quality aligned sentence pairs, but they may not fully represent the noise patterns, domains, and alignment errors found in large OPUS-derived corpora.
Moreover, this paper uses sentence-level retrieval because the target filtering unit is a sentence pair.
The results may not directly generalize to document-level or paragraph-level alignment settings, where discourse context and cross-sentence dependencies may affect parallelism.

The proposed routing framework relies on model scores as calibration signals rather than human-labeled filtering boundaries.
Direction-specific MRR, QE score distributions, margins, and coverage metadata are useful diagnostics, but they are still proxies for reliability.
For low-resource or low-confidence directions, the selected model may be the best available option without being strongly reliable in absolute terms.
Thresholds derived from benchmark or score distributions should therefore be interpreted cautiously, especially when benchmark evidence is sparse.

Our benchmarking covers a wide range of recent and representative models for both assessment components, but it is not an exhaustive list. Given the rapid development of multilingual embedding and QE systems, newer or alternative models may provide different performance profiles across the multilingual inventory, and our results represent a snapshot of the model landscape at the time of this study.

Finally, the coverage analysis relies on documented language support and language-code matching.
Such metadata does not prove that a model has seen a language during training, nor does it capture differences in script, dialect, register, or domain.
Coverage should therefore be understood as a practical reliability signal rather than a complete explanation of evaluator behavior.


\bibliography{references.bib}

\appendix

\section{QE Model Suite Details}
\label{app:qe-model-suite-details}

This appendix expands the reference-free QE model inventory summarized in
\Cref{tab:qe_models}.
The suite was selected to cover several practically distinct evaluator families:
dedicated encoder-based QE metrics, prompt-style learned metrics, decoder-only
LLM judges, reward-model evaluators, and a corpus-cleaning baseline.

\textsf{COMETKiwi} and \textsf{xCOMET} represent the encoder-based branch of
the suite.
Both are learned MT evaluation systems from the COMET family and use large
multilingual encoders to represent the source segment and the candidate
translation before predicting a sentence-level score
\citep{rei2023scaling_up_cometkiwi_wmt,guerreiro2024xcomet}.
\textsf{COMETKiwi} is a reference-free QE system with sentence-level and
word-level prediction components, whereas \textsf{xCOMET} extends this style of
metric with explicit error-span detection and can operate in reference-free,
reference-based, or combined modes.

\textsf{MetricX} represents a prompt-style learned metric rather than a free-form LLM judge.
The MetricX-24 family is initialized from multilingual T5-style encoder--decoder models and fine-tuned as a translation-quality regressor
\citep{juraska-etal-2024-metricx,xue2021mt5}.
In reference-free mode, the input contains the source and candidate translation,
and the model returns an MQM-style error score.
Because lower MetricX scores indicate better translation quality, we normalize MetricX by inverting its 0--25 scale before comparing it with the
other QE outputs.

\textsf{ReMedy} is included as a reward-model evaluator trained from human
preference comparisons for MT evaluation \citep{tan-monz-2025-remedy}.
Unlike direct regression metrics, ReMedy learns relative quality preferences and
then exposes a scalar score that can be used with or without references. In practice, the released framework included only the smaller set of languages emphasized in their published work, which
was insufficient for a FLORES-based experiment covering more than 200 language varieties. For this reason, the framework was patched locally so that its language-handling layer could accept a broader set of languages. This patch only expanded the framework’s language handling interface; it does not imply that ReMedy was trained on additional languages.

\textsf{M-Prometheus} and the three \textsf{Qwen3} variants are used as
decoder-only LLM judges.
\textsf{M-Prometheus} is an open multilingual judge trained on multilingual
direct-assessment and pairwise-comparison feedback \citep{pombal2025mprometheussuiteopenmultilingual}.
The \textsf{Qwen3} models are general open-weight multilingual LLMs
\citep{qwen3}.
For these LLM-based evaluators, we use the structured prompt in
\Cref{app:llm-prompt}, asking the model to produce seven dimension scores and
one overall reference-free quality score for each source--translation pair.

\textsf{Bicleaner} is included as a practical corpus-cleaning baseline rather
than as a dedicated MT metric.
Bicleaner AI is designed to identify noisy bitext by estimating whether two
sentences are mutual translations \citep{zaragoza2022bicleaner}.
It is therefore useful as an operational comparison point for corpus filtering,
even though its training objective is narrower than the explicit QE objectives
used by COMETKiwi, xCOMET, MetricX, ReMedy, and the LLM judges.

Table~\ref{tab:qe_model_suite_details} summarizes the model suite in terms of evaluator type, backbone architecture, approximate model size, and documented language coverage. For \textsf{ReMedy}, the documented language count refers to the released fine-tuned evaluator rather than to the possible multilingual capacity of its Gemma~2 backbone. Since Gemma~2 does not provide a precise public list of supported languages, the ReMedy coverage entry should be treated as documentation-based metadata rather than as a definitive estimate of the model's full language coverage.

\begin{table}[t]
  \centering
  \scriptsize
  \setlength{\tabcolsep}{4pt}
  \renewcommand{\arraystretch}{1.05}
  \begin{tabular}{@{}lllll@{}}
    \toprule
    Model & Model type & Backbone & Model size & Languages \\
    \midrule
    \textsf{COMETKiwi} & Encoder & XLM-R & $\sim$3.5B & 99 \\
    \textsf{xCOMET} & Encoder & XLM-R & $\sim$3.5B & 99 \\
    \textsf{MetricX} & Encoder--decoder & mT5 & $\sim$3.7B & 101 \\
    \textsf{ReMedy} & Decoder & Gemma 2 & $\sim$9B & 36 \\
    \textsf{M-Prometheus} & Decoder & Qwen2.5 & $\sim$7B & 30 \\
    \textsf{Qwen3-4B} & Decoder & -- & $\sim$4B & 119 \\
    \textsf{Qwen3-8B} & Decoder & -- & $\sim$8B & 119 \\
    \textsf{Qwen3-14B} & Decoder & -- & $\sim$14B & 119 \\
    \textsf{Bicleaner} & Encoder & XLM-R & $\sim$270M & 99 \\
    \bottomrule
  \end{tabular}
  \caption{
  Approximate QE model-suite summary.
  Backbone models, model sizes, and documented language counts come from model
  papers, model cards, or backbone documentation where available.
  The Languages column refers to the number of supported languages documented in model cards or technical reports.
  }
  \label{tab:qe_model_suite_details}
\end{table}

\section{Prompt Templates}
\label{app:prompt-templates}
This section presents the prompt templates utilized for the LLM-as-a-judge evaluations.

\subsection{Structured Batch Prompt}
\label{app:llm-prompt}

The following prompt template was used for the LLM-based evaluators.
The placeholders \texttt{\{batch\_size\}} and \texttt{\{items\_block\}} are populated dynamically at runtime with the number of segment pairs and their corresponding formatted source--translation content, respectively.

\begin{lstlisting}[style=prompttemplate]
You are a professional translation quality evaluator.

Below are {batch_size} source/translation segment pairs
to evaluate.

Source language: {source_lang}
Target language: {target_lang}

{items_block}

Task: Reference-free MT quality scoring for EVERY item
above.

Score each dimension as an integer 0..10 (higher = better),
then overall 0..100.

Dimensions:
1) accuracy_completeness
   (meaning preserved, no additions/omissions)
2) terminology_consistency
3) fluency_coherence
4) style_tone_audience
5) locale_formatting
   (numbers, punctuation, dates, tags if any)
6) technical_integrity
   (entities/units/code/markup preserved)
7) cultural_appropriateness

Output ONLY valid JSON with exactly this shape (no extra
keys, no text outside JSON, all values integers):

{
  "results": [
    {
      "id": <int>,
      "dims_0to10": {
        "accuracy_completeness": 0-10,
        "terminology_consistency": 0-10,
        "fluency_coherence": 0-10,
        "style_tone_audience": 0-10,
        "locale_formatting": 0-10,
        "technical_integrity": 0-10,
        "cultural_appropriateness": 0-10
      },
      "overall_0to100": 0-100
    }
  ]
}

Return exactly {batch_size} items in "results", one per
input segment, ordered by id.
\end{lstlisting}

\subsection{Simple Single-Segment Prompt}
\label{app:qwen-simple-prompt}

The following simpler prompt was used for the \textsf{Qwen3-4B} configuration in the prompt and batch-size sensitivity experiment.

\begin{lstlisting}[style=prompttemplate]
You are a professional translation quality evaluator.
Source language: {source_lang}
Target language: {target_lang}
Source text:
{source_seg}

Machine Translation text:
{target_seg}

Task: Reference-free MT quality scoring for this single segment.

Score the overall translation quality as an integer from 0 to 100
(higher = better).
Output ONLY valid JSON with exactly this shape (no extra keys,
no text outside JSON, value is an integer):
{"overall_0to100": 0-100}
\end{lstlisting}

\section{Detailed Results}
\label{app:detailed-results}

\subsection{Language Family-Level Results}
\label{sec:family-analysis}

Following the initial direction-level analysis in \Cref{sec:benchmark-single-model}, we investigate broader linguistic patterns within the results.
Given the substantial scale of 41{,}412 translation directions, an exhaustive qualitative discussion of each individual direction is analytically prohibitive.
Consequently, the analysis categorizes FLORES-200 languages according to Glottolog-based family assignments.\footnote{\url{https://glottolog.org/}}
This categorization yields 22 language families within the current dataset.
The distribution exhibits significant imbalance; Indo-European languages comprise 79 of the 204 varieties, followed by Atlantic-Congo~(34), Afro-Asiatic~(21), Austronesian~(21), Turkic~(11), and Sino-Tibetan~(9).
The remaining 16 families contribute between one and four varieties each, including Dravidian~(4), Tai-Kadai~(3), Uralic~(3), Nilotic~(3), Austroasiatic~(3), Mande~(2), Saharan~(2), and nine singleton families such as Kartvelian, Koreanic, and Japonic, as well as one artificial language.
To ensure analytical conciseness while maintaining representative cross-family variation, the qualitative family-level discussion is restricted to the four most prevalent families in FLORES-200: Indo-European, Atlantic-Congo, Afro-Asiatic, and Austronesian.

Figures~\ref{fig:family_scores_source} and~\ref{fig:family_scores_target} present a summary of the direction-level benchmark, categorized by top-level linguistic family using \textbf{Glottolog}'s comprehensive catalogue.
Each boxplot is constructed from family-pair mean scores rather than raw segment-level scores; thus, the dispersion reflects how the average direction-level performance of a model fluctuates across various family combinations.
The source-side panels in Figure~\ref{fig:family_scores_source} examine whether evaluator behavior varies according to the source language family, while the target-side panels in Figure~\ref{fig:family_scores_target} investigate the influence of the target language family.

\begin{figure*}[!tbp]
  \centering
  \begin{subfigure}{0.45\textwidth}
    \centering
    \includegraphics[
      width=\linewidth,
      alt={Boxplot of average QE scores when translating from Indo-European source languages to all target families. Nine models are shown on the x-axis, with average score on the y-axis; orange lines mark medians and black diamonds mark means. \textsf{MetricX}, \textsf{Qwen3-4B}, \textsf{Qwen3-8B}, and \textsf{ReMedy} have the highest central scores, while \textsf{Bicleaner}, COMETKiwi, and \textsf{xCOMET} are lower. Several LLM-based models show wide score ranges across target families.}
    ]{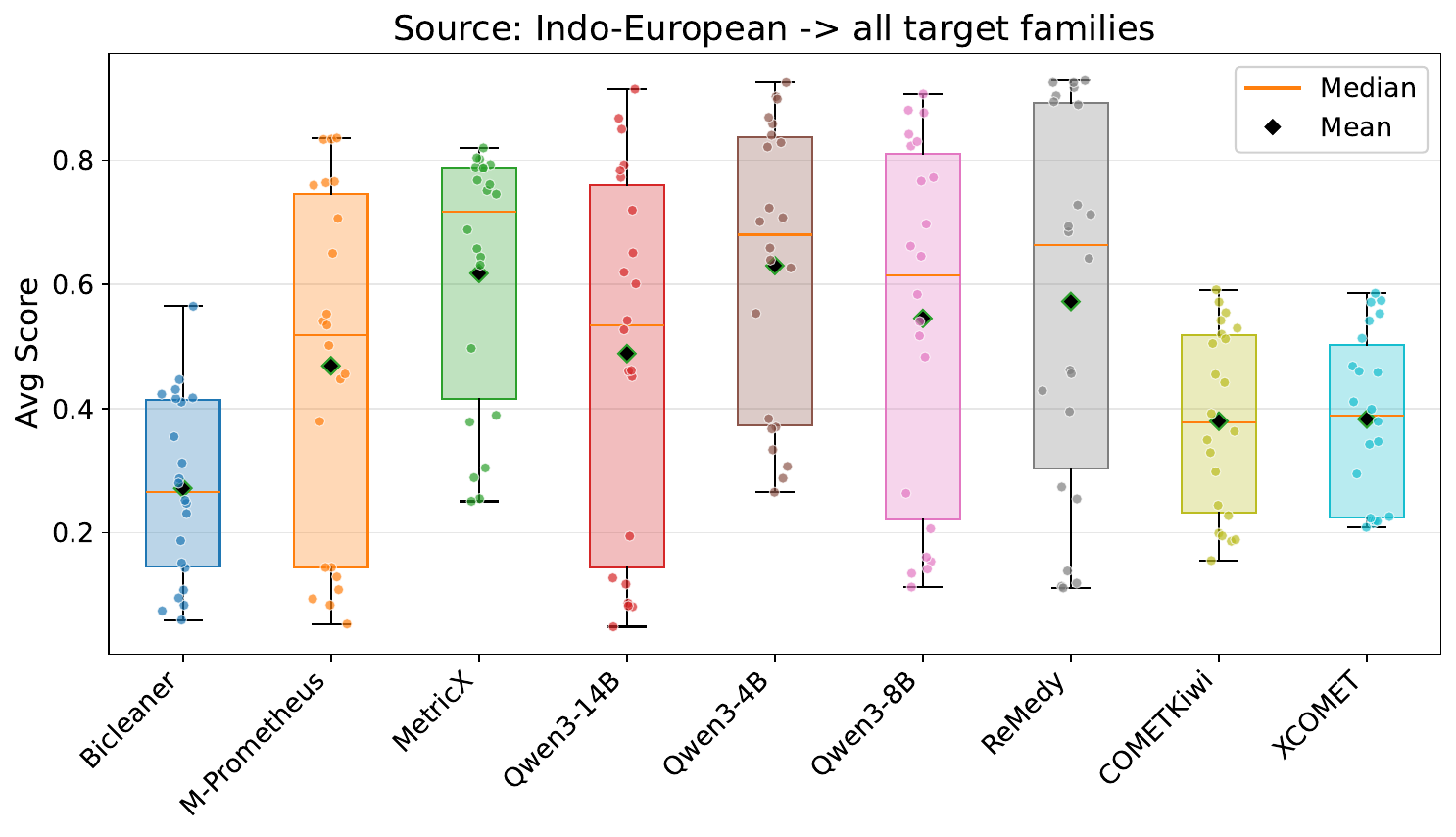}
    \caption{Indo-European source languages.}
  \end{subfigure}
  \quad
  \begin{subfigure}{0.45\textwidth}
    \centering
    \includegraphics[
      width=\linewidth,
      alt={Boxplot of average QE scores when translating from Atlantic-Congo source languages to all target families. Scores are generally lower than for Indo-European sources. \textsf{MetricX}, \textsf{Qwen3-4B}-2507, and \textsf{ReMedy} remain among the stronger models, while \textsf{Bicleaner} is the lowest and \textsf{xCOMET} is low with a narrow spread. The LLM-based models show substantial variation across target families.}
    ]{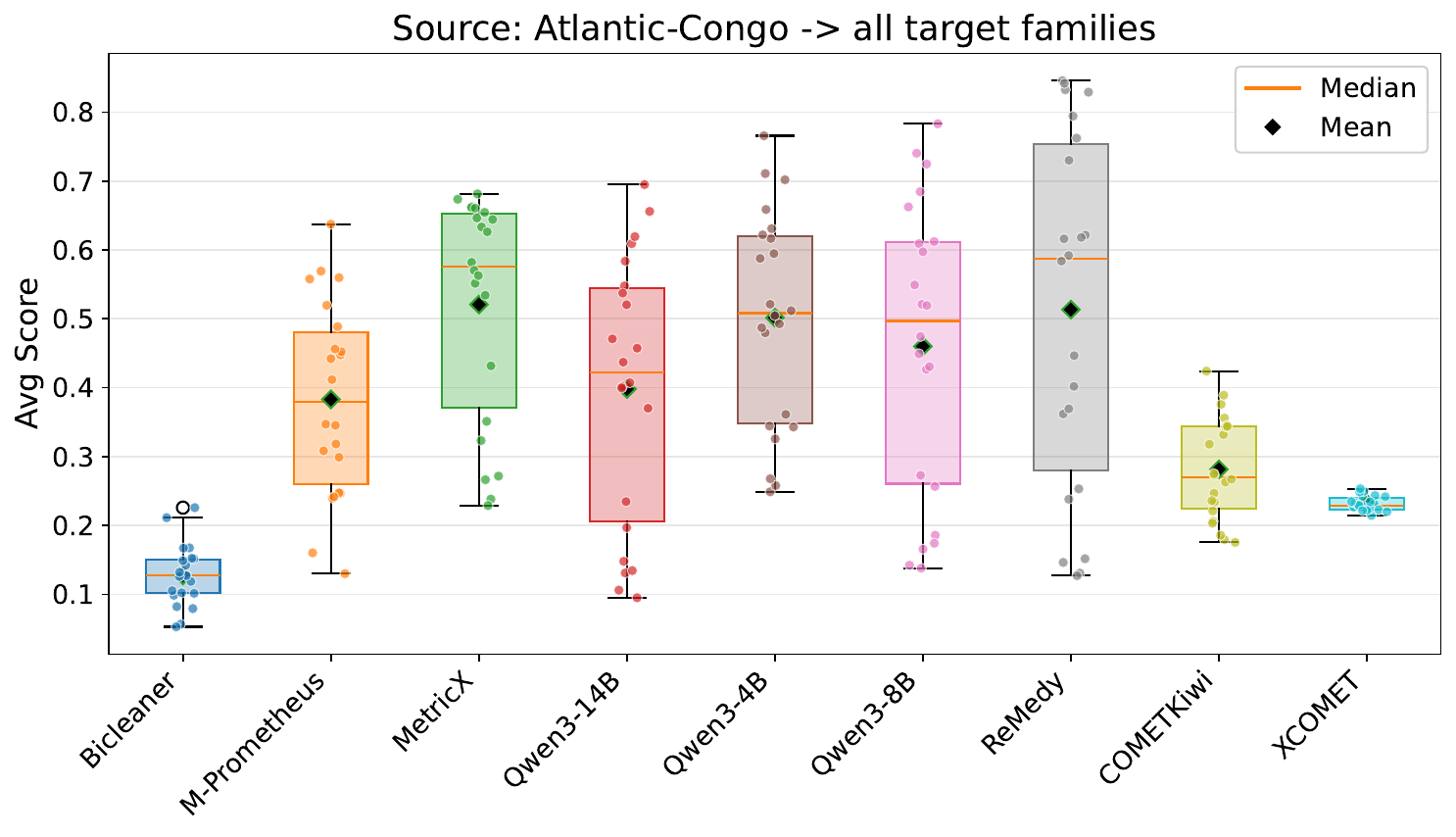}
    \caption{Atlantic-Congo source languages.}
  \end{subfigure}
  \caption{Source-side family comparison for the four largest Glottolog families. For each source family, each box summarizes one model's family-pair mean scores over all target families.}
  \label{fig:family_scores_source}
\end{figure*}

\begin{figure*}[!tbp]
  \ContinuedFloat
  \centering
  \begin{subfigure}{0.45\textwidth}
    \centering
    \includegraphics[
      width=\linewidth,
      alt={Boxplot of average QE scores when translating from Afro-Asiatic source languages to all target families. \textsf{MetricX}, \textsf{Qwen3-4B}-2507, and \textsf{ReMedy} have the strongest central scores. \textsf{Qwen3-14B} and \textsf{Qwen3-8B} have wide spreads, indicating that their scores depend strongly on the target family. \textsf{Bicleaner}, COMETKiwi, and \textsf{xCOMET} are lower overall.}
    ]{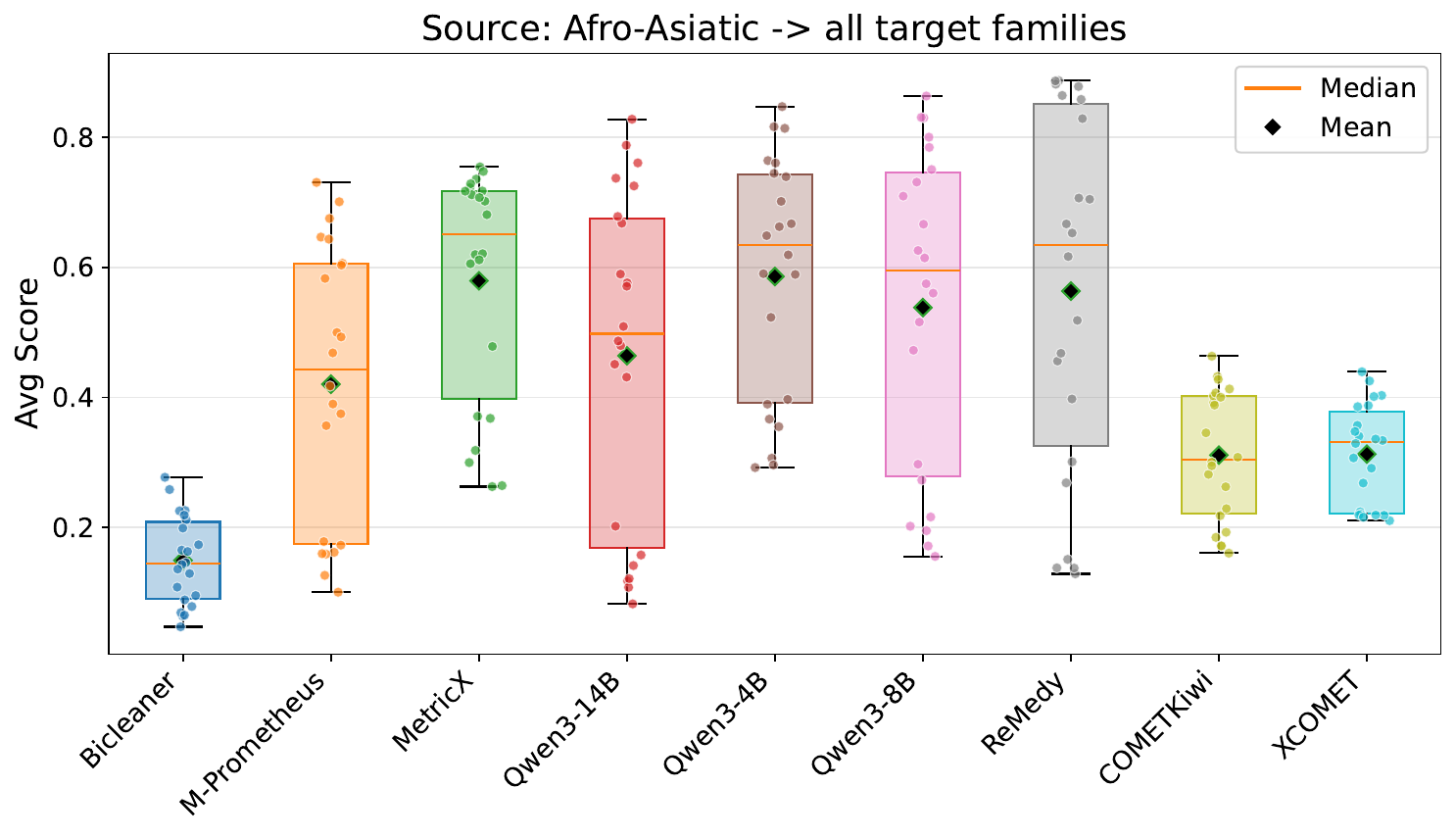}
    \caption{Afro-Asiatic source languages.}
  \end{subfigure}
  \quad
  \begin{subfigure}{0.45\textwidth}
    \centering
    \includegraphics[
      width=\linewidth,
      alt={Boxplot of average QE scores when translating from Austronesian source languages to all target families. \textsf{MetricX}, \textsf{Qwen3-4B}-2507, and \textsf{ReMedy} are again among the strongest models, while \textsf{Bicleaner} has the lowest central scores. M-Prometheus, \textsf{Qwen3-14B}, \textsf{Qwen3-8B}, and \textsf{ReMedy} show broad score ranges across target families.}
    ]{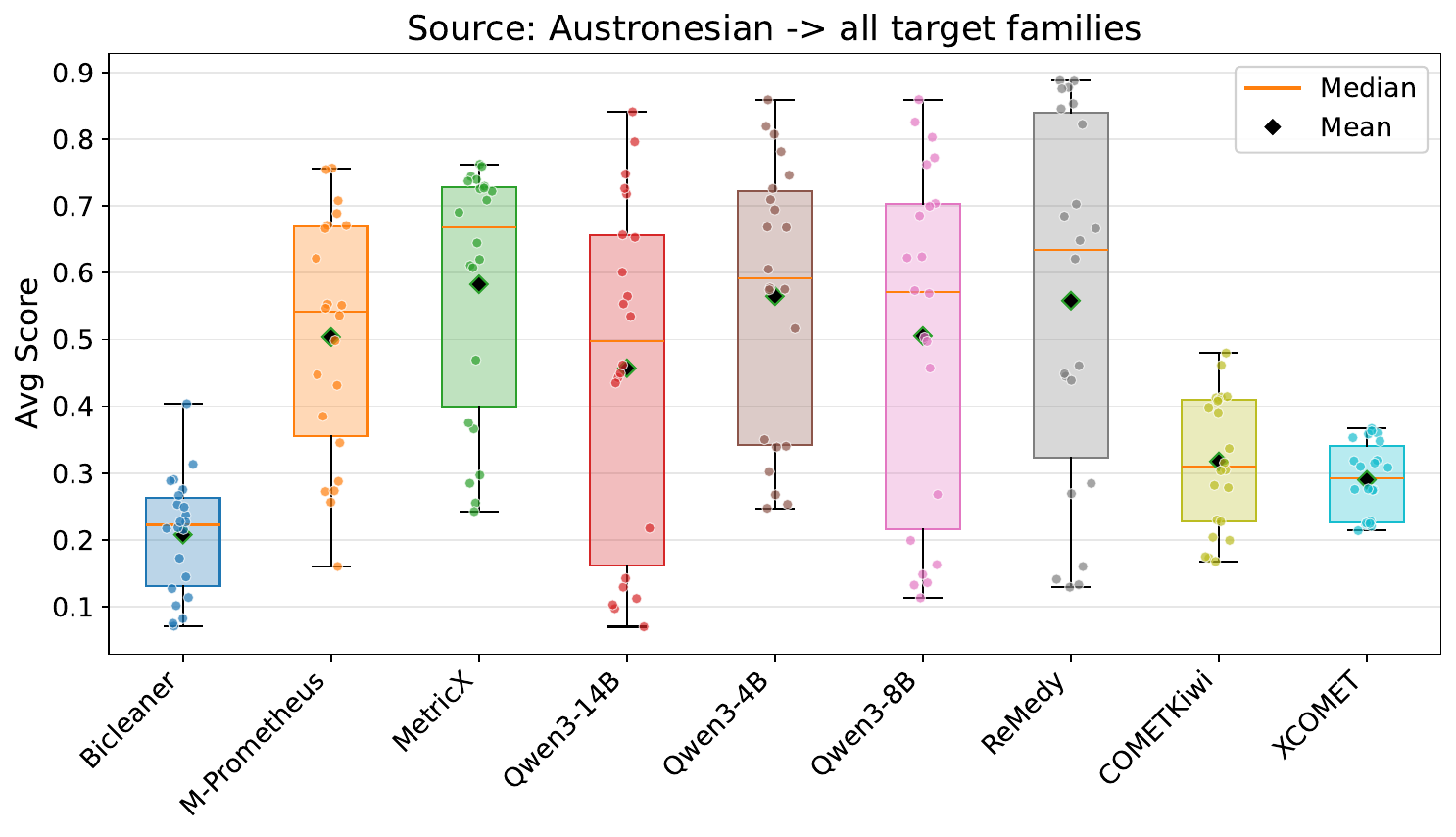}
    \caption{Austronesian source languages.}
  \end{subfigure}
  \caption{Source-side family comparison for the four largest Glottolog families, continued.}
\end{figure*}

\begin{figure*}[!tbp]
  \centering
  \begin{subfigure}{0.45\textwidth}
    \centering
    \includegraphics[
      width=\linewidth,
      alt={Boxplot of average QE scores when translating from all source families into Indo-European target languages. The leading models have high central scores: \textsf{Qwen3-4B}-2507, \textsf{Qwen3-8B}, \textsf{MetricX}, and \textsf{ReMedy} cluster near the top. \textsf{Bicleaner} is much lower, while COMETKiwi and \textsf{xCOMET} sit in the lower-middle range. The target-side distributions are comparatively compact for several leading models.}
    ]{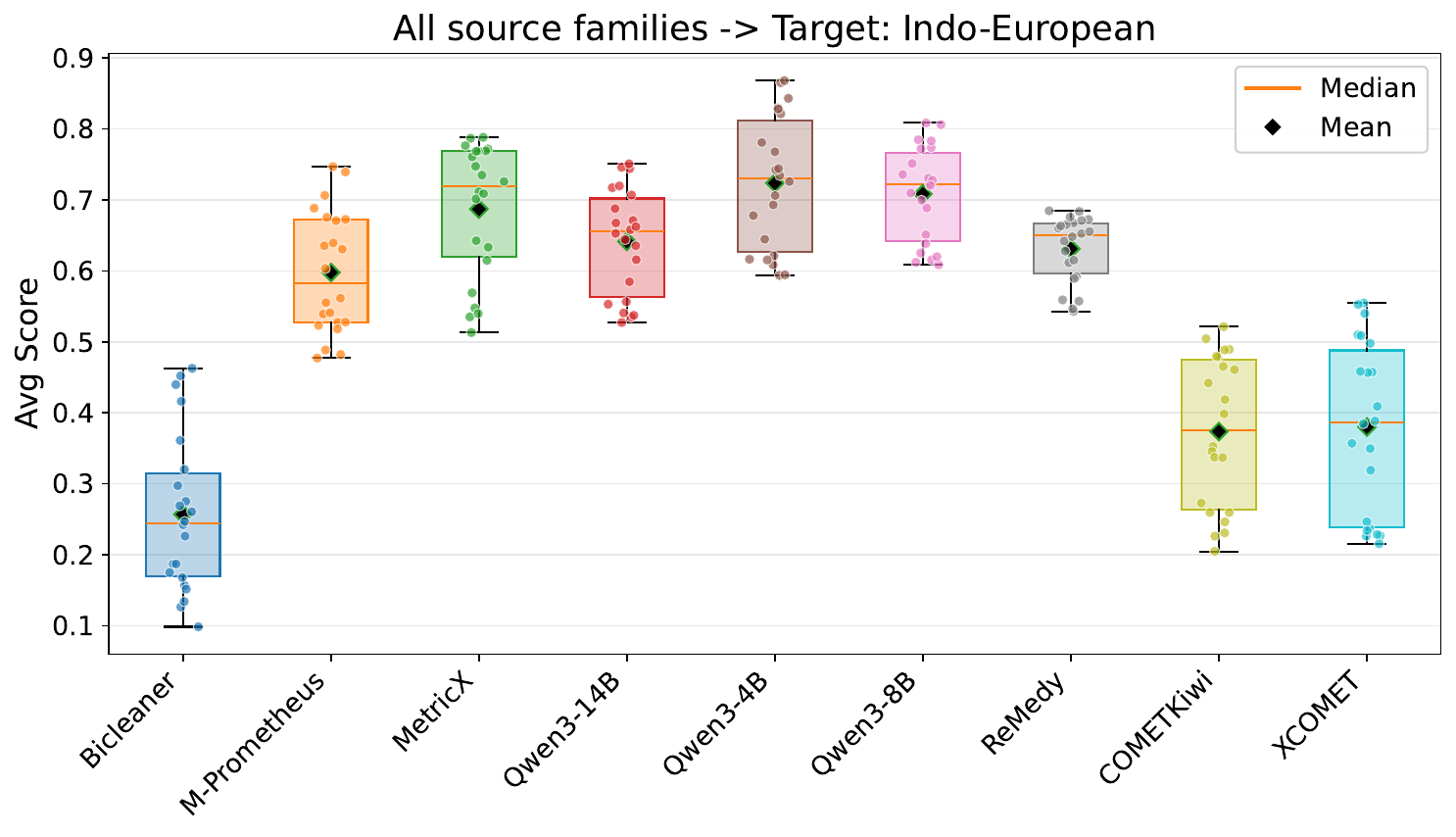}
    \caption{Indo-European target languages.}
  \end{subfigure}
  \quad
  \begin{subfigure}{0.45\textwidth}
    \centering
    \includegraphics[
      width=\linewidth,
      alt={Boxplot of average QE scores when translating from all source families into Atlantic-Congo target languages. This target family has the lowest overall score range. \textsf{MetricX} and \textsf{ReMedy} are the strongest relative performers, \textsf{Qwen3-4B}-2507 is lower but still competitive, and \textsf{Bicleaner} is the lowest. Most models have medians below 0.5.}
    ]{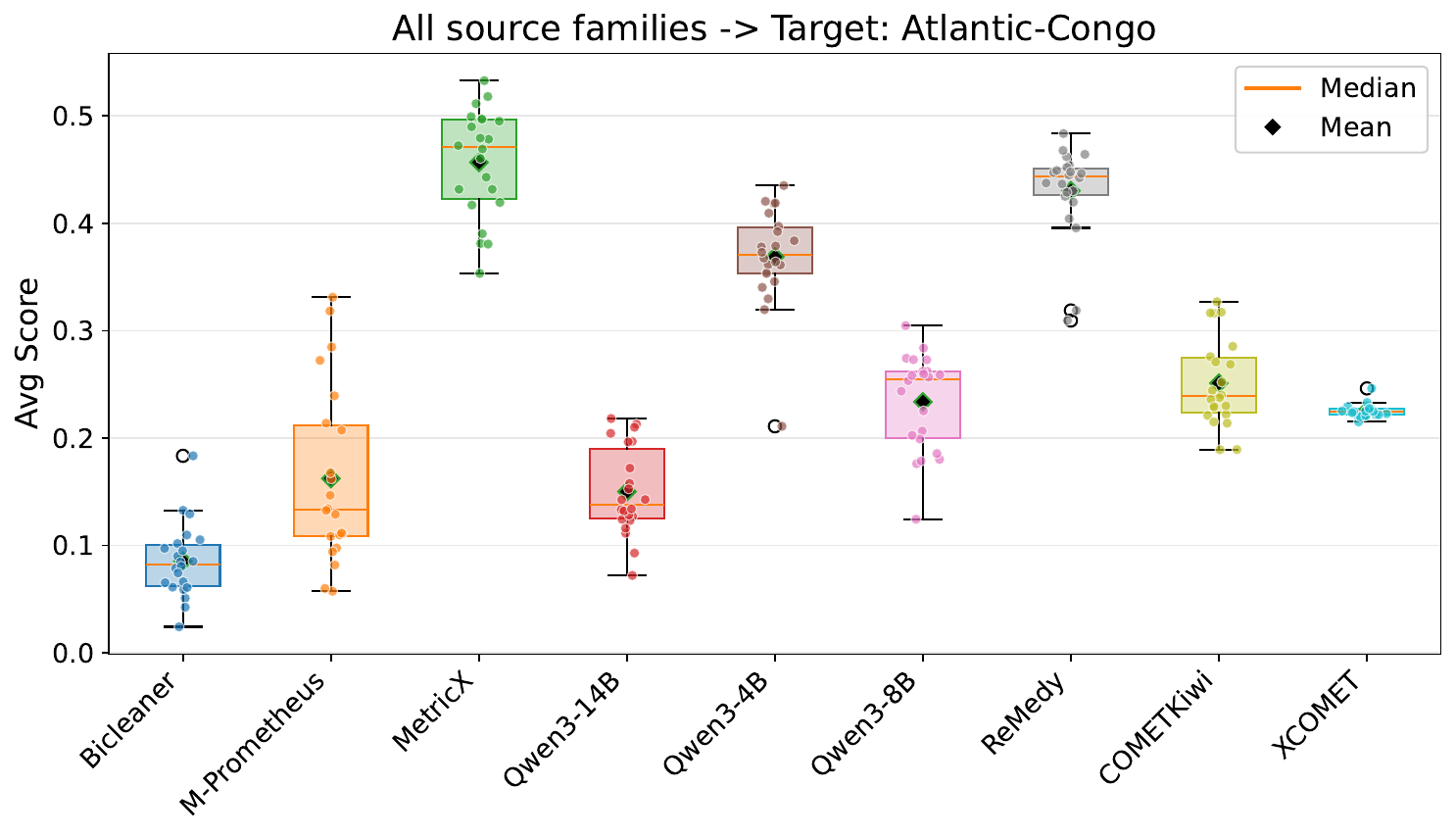}
    \caption{Atlantic-Congo target languages.}
  \end{subfigure}
  \caption{Target-side family comparison for the four largest Glottolog families. For each target family, each box summarizes one model's family-pair mean scores over all source families.}
  \label{fig:family_scores_target}
\end{figure*}

\begin{figure*}[!tbp]
  \ContinuedFloat
  \centering
  \begin{subfigure}{0.45\textwidth}
    \centering
    \includegraphics[
      width=\linewidth,
      alt={Boxplot of average QE scores when translating from all source families into Afro-Asiatic target languages. \textsf{MetricX} and \textsf{Qwen3-4B}-2507 have the highest central scores, followed by \textsf{Qwen3-8B} and M-Prometheus. \textsf{ReMedy} is lower than in the source-side plots, and \textsf{Bicleaner} has the lowest scores.}
    ]{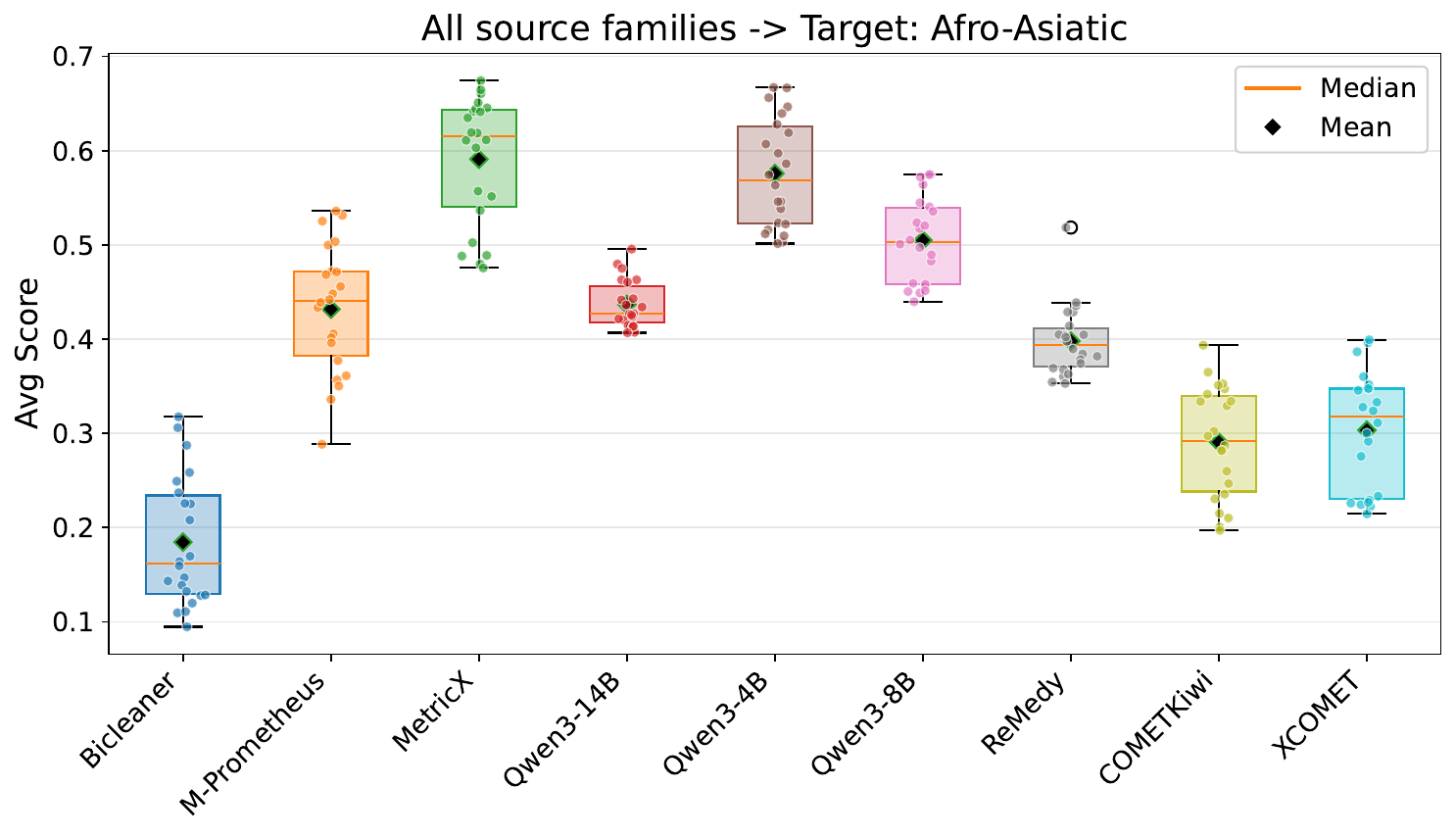}
    \caption{Afro-Asiatic target languages.}
  \end{subfigure}
  \quad
  \begin{subfigure}{0.45\textwidth}
    \centering
    \includegraphics[
      width=\linewidth,
      alt={Boxplot of average QE scores when translating from all source families into Austronesian target languages. \textsf{MetricX} and \textsf{Qwen3-4B}-2507 have the highest central scores, with \textsf{Qwen3-8B} and M-Prometheus in the middle. \textsf{ReMedy} is lower than the leading models in this target-family view, while \textsf{Bicleaner} remains the lowest.}
    ]{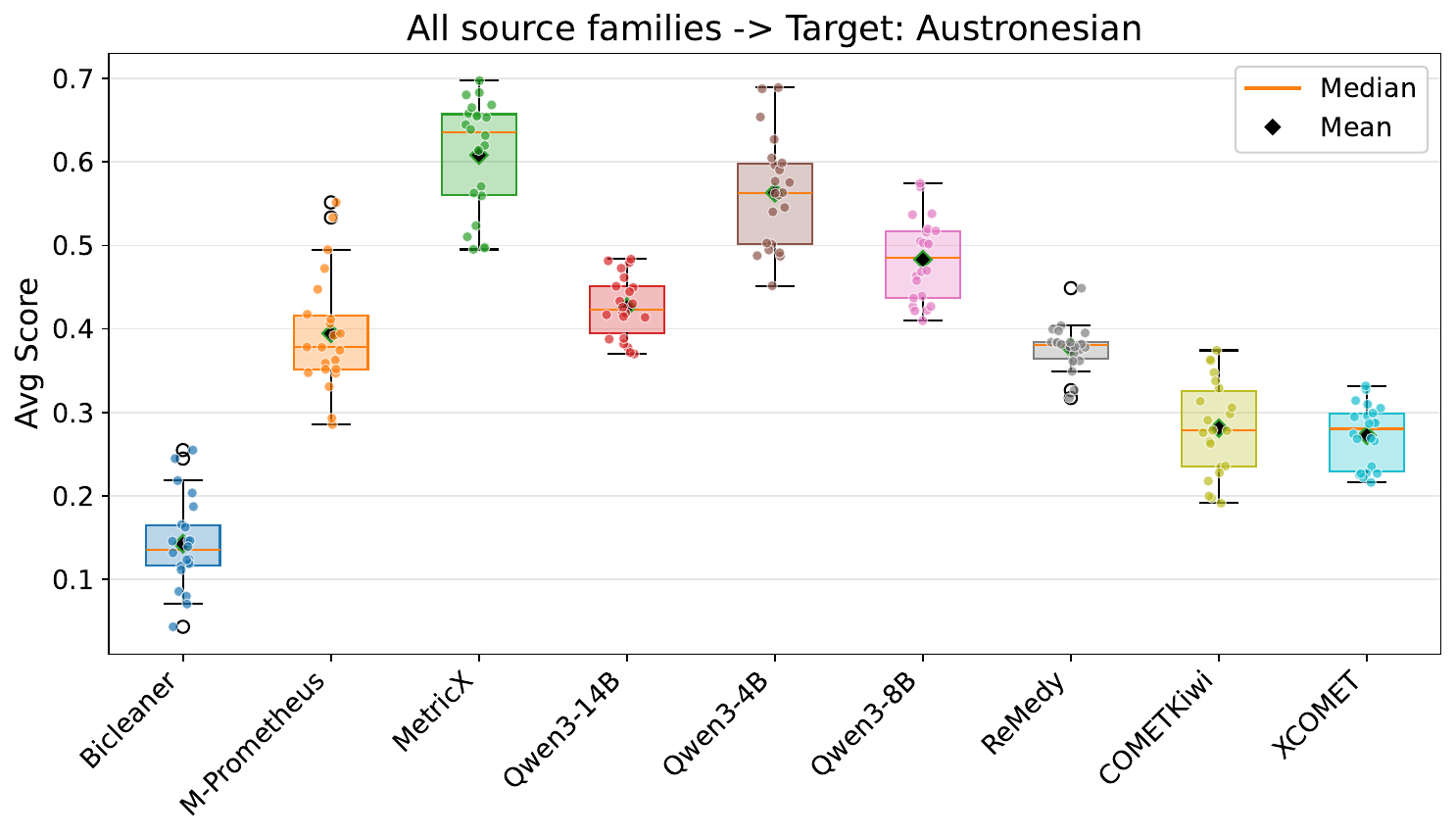}
    \caption{Austronesian target languages.}
  \end{subfigure}
  \caption{Target-side family comparison for the four largest Glottolog families, continued.}
\end{figure*}

Within this descriptive framework, Figure~\ref{fig:family_scores_source} indicates that the aggregate \textbf{RQ1} ranking is not uniformly influenced by all source-language groupings.
Indo-European source directions demonstrate a general upward shift among higher-performing evaluators, specifically \textsf{MetricX}, \textsf{Qwen3-4B}, \textsf{Qwen3-8B}, and \textsf{ReMedy}.
Austronesian and Afro-Asiatic source languages exhibit a comparable intermediate range, although the relative ordering of individual models fluctuates between the two panels.
Performance on Atlantic-Congo source directions is consistently lower across the majority of models, most notably for \textsf{Bicleaner}, \textsf{COMETKiwi}, \textsf{xCOMET}, \textsf{M-Prometheus}, and the Qwen variants.
This trend is consistent with a resource-availability hypothesis, as Indo-European languages are more extensively represented in large-scale pretraining corpora than many Atlantic-Congo languages; this asymmetry may subsequently impact the representations leveraged by reference-free evaluators.
Model-level score dispersion further contextualizes the win-count results: \textsf{ReMedy}, \textsf{Qwen3-14B}, and \textsf{Qwen3-8B} exhibit substantial source-side variance, whereas \textsf{MetricX} maintains relatively stable high central scores across all four source families.
Consequently, the source-family analysis supports the interpretation that \textsf{ReMedy}'s direction-level superiority is partially attributable to favorable family pairings rather than consistent cross-family performance.

The target-side panels in Figure~\ref{fig:family_scores_target} demonstrate a more pronounced family effect than the source-side equivalents.
Relative to the source-side visualizations, they exhibit clearer shifts in central scores by family and, for numerous models, reduced within-family dispersion.
Indo-European target languages yield the highest central scores for leading evaluators, with \textsf{Qwen3-4B}, \textsf{Qwen3-8B}, and \textsf{MetricX} clustering near the upper bound, while lower-performing models also demonstrate an upward shift compared to other target families.
Atlantic-Congo targets represent the lowest target-side grouping; most models experience a substantial performance decline, although \textsf{MetricX} and \textsf{ReMedy} remain the most competitive systems within this subset, despite mean and median scores remaining below 0.5.
Afro-Asiatic and Austronesian targets generally occupy an intermediate position, with \textsf{MetricX} and \textsf{Qwen3-4B} maintaining the highest central scores.

The relative compactness of many target-side distributions suggests that evaluator scores are more consistent once the target language family is held constant. This pattern indicates that target-side properties may explain more of the observed evaluator behaviour than source-side properties.
This observation is plausible given that reference-free evaluators must assess the fluency and acceptability of the translated segment in the target language, a task that depends directly on the model's internal proficiency in that language.
This asymmetry is also explored in the coverage-aware analysis in \Cref{sec:benchmark-coverage}.

\subsection{Results of Ensemble-based Methods}

Table~\ref{tab:ensemble_benchmark_full} provides the full RQ2 comparison underlying the compact ensemble summary in \Cref{sec:benchmark-ensembles}.
It adds nine ensemble configurations to the nine individual evaluators, so the rank statistics are computed over an expanded 18-method pool and therefore differ from the single-model ranks in Table~\ref{tab:single_model_results}.
Because some ensemble variants are defined only when their constituent pool is non-empty, the table uses two denominators: \textbf{Win~\%} is normalized by the full 41{,}412-direction benchmark, whereas \textbf{Macro}, \textbf{Last}, and \textbf{Rank} are computed over the method's eligible directions.

The ensemble names encode both the aggregation rule and the coverage filter.
\texttt{mean}, \texttt{median}, and \texttt{wavg} denote mean aggregation, median aggregation, and weighted averaging, respectively.

The coverage qualifier specifies how constituent models are selected for each direction. \texttt{Both} includes only constituent models that document support for both the source and target languages, \texttt{Source-seen} includes only models that document source-language support, and \texttt{Target-seen} includes only models that document target-language support. Ensemble names without a coverage qualifier denote full-pool ensembles, which apply no coverage-based eligibility restriction and therefore include all 41{,}412 directions.

The table supplements the RQ2 discussion by reporting the complete method list, last-place counts, largest-margin wins, and eligible-direction counts omitted from the compact table in the main text.

\begin{table*}[t]
  \centering
  \scriptsize
  \setlength{\tabcolsep}{3pt}
    \resizebox{\textwidth}{!}{
  \renewcommand{\arraystretch}{0.95}
  \begin{tabular}{@{}lcrrccr@{}}
    \toprule
    Method & Wins $\uparrow$ & Macro $\uparrow$ & Last $\downarrow$ & Rank $\downarrow$ & Margin $\uparrow$ & N \\
    \midrule
    \textsf{ReMedy}
      & 16{,}306 (39.4) & 0.5489 & 361 & $5.69{\pm}5.34$ & 0.4032 & 41{,}412 \\

    \textsf{Qwen3-4B}
      & 11{,}991 (29.0) & 0.6160 & 0 & $2.77{\pm}1.98$ & 0.2791 & 41{,}412 \\

    \textsf{MetricX}
      & 8{,}771 (21.2) & 0.6228 & 0 & $4.17{\pm}3.19$ & 0.4126 & 41{,}412 \\

    \textsf{M-Prometheus}
      & 1{,}815 (4.4) & 0.4751 & 1{,}681 & $8.22{\pm}4.16$ & 0.4961 & 41{,}412 \\

    \textsf{Qwen3-8B}
      & 1{,}679 (4.1) & 0.5517 & 29 & $5.37{\pm}3.08$ & 0.1977 & 41{,}412 \\

    \textsf{Qwen3-14B}
      & 460 (1.1) & 0.4879 & 741 & $8.03{\pm}3.60$ & 0.0836 & 41{,}412 \\

    \textsf{Bicleaner}
      & 206 (0.5) & 0.2141 & 7{,}639 & $13.33{\pm}2.70$ & 0.3592 & 41{,}412 \\

    \textsf{COMETKiwi}
      & 38 (0.1) & 0.3284 & 678 & $11.10{\pm}3.63$ & 0.0660 & 41{,}412 \\

    \textsf{xCOMET}
      & 22 (0.1) & 0.3221 & 406 & $11.04{\pm}3.73$ & 0.0325 & 41{,}412 \\
      
    \midrule

    \textsf{Both-unseen weighted ens.}
      & 60 (0.1) & 0.4526 & 1 & $6.90{\pm}4.13$ & 0.0000 & 30{,}588 \\

    \textsf{Both-seen median ens.}
      & 35 (0.1) & 0.7135 & 0 & $6.40{\pm}3.27$ & 0.0050 & 19{,}744 \\

    \textsf{Both-seen weighted ens.}
      & 13 (0.0) & 0.7179 & 0 & $6.51{\pm}3.37$ & 0.0000 & 19{,}744 \\

    \textsf{Source-seen weighted ens.}
      & 11 (0.0) & 0.5504 & 0 & $6.70{\pm}3.02$ & 0.0000 & 29{,}841 \\

    \textsf{Target-seen weighted ens.}
      & 5 (0.0) & 0.6417 & 0 & $6.99{\pm}3.28$ & 0.0000 & 29{,}841 \\

    \textsf{Mean ens.}
      & 0 (0.0) & 0.4630 & 0 & $9.08{\pm}2.50$ & -- & 41{,}412 \\

    \textsf{Median ens.}
      & 0 (0.0) & 0.4842 & 0 & $8.09{\pm}1.71$ & -- & 41{,}412 \\

    \textsf{Both-seen mean ens.}
      & 0 (0.0) & 0.6901 & 0 & $8.20{\pm}4.09$ & -- & 19{,}744 \\

    \textsf{Weighted ens.}
      & 0 (0.0) & 0.5026 & 0 & $7.03{\pm}2.27$ & -- & 41{,}412 \\

    \bottomrule
  \end{tabular}
  }
  \caption{
  Full comparison of individual QE evaluators and unsupervised ensemble configurations for RQ2.
  Wins report count and percentage.
  Last denotes the number of directions where a method ranks last.
  Rank reports mean $\pm$ standard deviation.
  Margin is the largest observed benchmark-margin victory.
  N denotes eligible translation directions.
  Rank statistics are computed in the expanded pool containing both individual evaluators and ensemble variants.
  }
  \label{tab:ensemble_benchmark_full}
\end{table*}

\subsection{Language Coverage Analysis}
\label{app:coverage-analysis}

Tables~\ref{tab:coverage_analysis_models} and~\ref{tab:coverage_analysis_ensembles} provide the detailed coverage-stratified statistics behind the RQ3 discussion in \Cref{sec:benchmark-coverage}.
The single-model table groups each evaluator by documented source-language support, target-language support, and joint source--target support.
The ensemble table is a separate coverage diagnostic: unlike the full-pool RQ2 ensembles in Table~\ref{tab:ensemble_benchmark_full}, it uses the reduced coverage-analysis pool consisting of \textsf{MetricX}, \textsf{ReMedy}, \textsf{M-Prometheus}, and the \textsf{Qwen3} variants.
This difference explains why the all-row ensemble macro-averages in the coverage table are not identical to the unrestricted ensemble scores in the RQ2 table. We execluded CoMETKiwi, xCOMET, and Bicleaner from the coverage analysis due to their low overall performance in the current benchmark.

\begin{table*}[t]
  \centering
  \scriptsize
  \setlength{\tabcolsep}{3pt}
  \renewcommand{\arraystretch}{0.92}
  \resizebox{\textwidth}{!}{
  \begin{tabular}{@{}llcrrccr@{}}
    \toprule
    Method & Cov. & Wins $\uparrow$ & Macro $\uparrow$ & Rank $\downarrow$ & Margin $\uparrow$ & Top-1/3 $\uparrow$ & N \\
    \midrule
    \textsf{ReMedy} & All & 16{,}133 (39.0) & 0.5489 & $5.77{\pm}5.26$ & 0.3927 & 39.0 / 54.1 & 41{,}412 \\
     & Source & 2{,}910 (32.6) & 0.5728 & $5.97{\pm}5.23$ & 0.3802 & 32.6 / 51.9 & 8{,}932 \\
     & Target & 4{,}393 (49.2) & 0.7354 & $4.39{\pm}4.73$ & 0.2602 & 49.2 / 66.5 & 8{,}932 \\
     & Both & 665 (35.1) & 0.7803 & $4.61{\pm}4.53$ & 0.1255 & 35.1 / 61.9 & 1{,}892 \\
    \midrule

    \textsf{Qwen3-4B} & All & 11{,}986 (28.9) & 0.6160 & $3.34{\pm}2.73$ & 0.2791 & 28.9 / 68.9 & 41{,}412 \\
     & Source & 9{,}316 (37.6) & 0.6726 & $2.77{\pm}2.48$ & 0.2791 & 37.6 / 78.1 & 24{,}766 \\
     & Target & 9{,}152 (37.0) & 0.7691 & $3.33{\pm}3.01$ & 0.2275 & 37.0 / 69.7 & 24{,}766 \\
     & Both & 7{,}532 (51.0) & 0.8498 & $2.45{\pm}2.58$ & 0.2275 & 51.0 / 84.1 & 14{,}762 \\
    \midrule

    \textsf{MetricX} & All & 8{,}668 (20.9) & 0.6228 & $5.42{\pm}4.31$ & 0.3875 & 20.9 / 51.5 & 41{,}412 \\
     & Source & 5{,}082 (23.0) & 0.6657 & $5.43{\pm}4.48$ & 0.3867 & 23.0 / 53.4 & 22{,}127 \\
     & Target & 2{,}180 (9.9) & 0.7233 & $7.06{\pm}4.32$ & 0.2189 & 9.9 / 33.8 & 22{,}127 \\
     & Both & 1{,}182 (10.0) & 0.7776 & $7.22{\pm}4.48$ & 0.1378 & 10.0 / 34.7 & 11{,}772 \\
    \midrule

    \textsf{M-Prometheus} & All & 1{,}815 (4.4) & 0.4751 & $9.21{\pm}3.86$ & 0.4961 & 4.4 / 15.3 & 41{,}412 \\
     & Source & 127 (1.5) & 0.5274 & $9.31{\pm}3.48$ & 0.2196 & 1.5 / 12.1 & 8{,}323 \\
     & Target & 208 (2.5) & 0.7165 & $8.92{\pm}3.26$ & 0.2140 & 2.5 / 10.7 & 8{,}323 \\
     & Both & 12 (0.7) & 0.8132 & $8.94{\pm}2.72$ & 0.1054 & 0.7 / 5.6 & 1{,}640 \\
    \midrule

    \textsf{Qwen3-8B} & All & 1{,}676 (4.0) & 0.5517 & $6.34{\pm}3.25$ & 0.1865 & 4.0 / 27.1 & 41{,}412 \\
     & Source & 328 (1.3) & 0.5858 & $6.86{\pm}3.22$ & 0.0821 & 1.3 / 20.9 & 24{,}766 \\
     & Target & 1{,}619 (6.5) & 0.7466 & $5.19{\pm}3.17$ & 0.1865 & 6.5 / 40.1 & 24{,}766 \\
     & Both & 327 (2.2) & 0.8029 & $5.88{\pm}3.25$ & 0.0821 & 2.2 / 31.0 & 14{,}762 \\
    \midrule

    \textsf{Qwen3-14B} & All & 456 (1.1) & 0.4879 & $9.41{\pm}3.17$ & 0.0763 & 1.1 / 7.6 & 41{,}412 \\
     & Source & 343 (1.4) & 0.5191 & $10.05{\pm}3.24$ & 0.0432 & 1.4 / 7.4 & 24{,}766 \\
     & Target & 456 (1.8) & 0.6981 & $8.31{\pm}3.41$ & 0.0763 & 1.8 / 12.5 & 24{,}766 \\
     & Both & 343 (2.3) & 0.7620 & $8.96{\pm}3.62$ & 0.0432 & 2.3 / 12.3 & 14{,}762 \\
    \midrule

    \textsf{Bicleaner} & All & 206 (0.5) & 0.2141 & $13.07{\pm}2.57$ & 0.3592 & 0.5 / 1.8 & 41{,}412 \\
     & Source & 52 (0.3) & 0.2596 & $13.25{\pm}2.70$ & 0.2256 & 0.3 / 1.6 & 20{,}503 \\
     & Target & 30 (0.1) & 0.3014 & $13.66{\pm}2.18$ & 0.0877 & 0.1 / 0.9 & 20{,}503 \\
     & Both & 25 (0.2) & 0.3910 & $13.85{\pm}2.55$ & 0.0877 & 0.2 / 1.6 & 10{,}100 \\
    \midrule

    \textsf{COMETKiwi} & All & 38 (0.1) & 0.3284 & $11.06{\pm}3.51$ & 0.0660 & 0.1 / 4.9 & 41{,}412 \\
     & Source & 24 (0.1) & 0.3956 & $10.86{\pm}4.01$ & 0.0353 & 0.1 / 7.7 & 20{,}503 \\
     & Target & 0 (0.0) & 0.4341 & $12.35{\pm}2.56$ & 0.0000 & 0.0 / 0.8 & 20{,}503 \\
     & Both & 0 (0.0) & 0.5354 & $12.75{\pm}2.79$ & 0.0000 & 0.0 / 1.3 & 10{,}100 \\
    \midrule

    \textsf{xCOMET} & All & 22 (0.1) & 0.3221 & $10.98{\pm}3.53$ & 0.0325 & 0.1 / 5.2 & 41{,}412 \\
     & Source & 14 (0.1) & 0.3932 & $10.92{\pm}3.67$ & 0.0325 & 0.1 / 5.0 & 20{,}503 \\
     & Target & 10 (0.0) & 0.4148 & $12.66{\pm}2.19$ & 0.0325 & 0.0 / 0.3 & 20{,}503 \\
     & Both & 10 (0.1) & 0.5494 & $12.52{\pm}2.40$ & 0.0325 & 0.1 / 0.6 & 10{,}100 \\
    \bottomrule
  \end{tabular}
  }
  \caption{
  Coverage-restricted single-model results for RQ3.
  Cov. denotes coverage subset.
  Wins report count and percentage.
  Rank reports mean $\pm$ standard deviation.
  Top-1/3 reports Top-1\% and Top-3\%.
  N denotes eligible directions.
  }
  \label{tab:coverage_analysis_models}
\end{table*}

\begin{table*}[t]
  \centering
  \scriptsize
  \setlength{\tabcolsep}{3pt}
  \renewcommand{\arraystretch}{0.95}
  \resizebox{\textwidth}{!}{
  \begin{tabular}{@{}llcrrccr@{}}
    \toprule
    Ensemble & Cov. & Wins $\uparrow$ & Macro $\uparrow$ & Rank $\downarrow$ & Margin $\uparrow$ & Top-1/3 $\uparrow$ & N \\
    \midrule
    \texttt{mean} & All & 0 (0.0) & 0.5504 & $7.05{\pm}1.97$ & 0.0000 & 0.0 / 0.5 & 41{,}412 \\
    \texttt{median} & All & 0 (0.0) & 0.5531 & $5.96{\pm}2.10$ & 0.0000 & 0.0 / 12.8 & 41{,}412 \\
    \texttt{wavg} & All & 0 (0.0) & 0.5562 & $6.19{\pm}2.06$ & 0.0000 & 0.0 / 5.9 & 41{,}412 \\
    \midrule
    \texttt{mean} & Both & 0 (0.0) & 0.7817 & $6.21{\pm}3.05$ & 0.0000 & 5.3 / 21.9 & 19{,}058 \\
    \texttt{median} & Both & 48 (0.3) & 0.7864 & $4.89{\pm}2.72$ & 0.0050 & 5.6 / 37.1 & 19{,}058 \\
    \texttt{wavg} & Both & 63 (0.3) & 0.7837 & $5.57{\pm}2.95$ & 0.0000 & 5.4 / 31.5 & 19{,}058 \\
    \midrule
    \texttt{wavg} & Source & 45 (0.2) & 0.6069 & $5.80{\pm}2.81$ & 0.0000 & 4.5 / 24.3 & 29{,}232 \\
    \texttt{wavg} & Target & 23 (0.1) & 0.7175 & $5.84{\pm}2.97$ & 0.0000 & 4.6 / 28.2 & 29{,}232 \\
    \texttt{wavg} & Neither & 233 (0.8) & 0.4751 & $6.95{\pm}4.10$ & 0.0000 & 5.3 / 29.5 & 30{,}384 \\
    \bottomrule
  \end{tabular}
  }
  \caption{
  Coverage-restricted ensemble results for RQ3, computed with the reduced coverage-analysis evaluator pool.
  Cov. denotes coverage subset.
  Wins report count and percentage.
  Rank reports mean $\pm$ standard deviation.
  Top-1/3 reports Top-1\% and Top-3\%.
  N denotes eligible directions.
  }
  \label{tab:coverage_analysis_ensembles}
\end{table*}

The \textit{Coverage subset} column in \Cref{tab:coverage_analysis_models,tab:coverage_analysis_ensembles} specifies the subset of directions used for each row.
For single evaluators, \textit{Source}, \textit{Target}, and \textit{Both} denote directions where the source language, target language, or both languages are documented as supported; these subsets overlap and therefore should not be summed.
For ensembles, the same labels denote the coverage condition used to select constituent evaluators for each direction, and rows include only directions with at least one eligible constituent model.
\textit{Neither} denotes the ensemble condition in which only constituent evaluators with no documented support for either side of the direction are included.
Win count and Win~\% summarize first-place finishes within the eligible subset, while Top-1~\% and Top-3~\% indicate how often a method ranks first or among the top three candidates, including ties.
The detailed rows are intended to support the body-level interpretation in \Cref{sec:benchmark-coverage}.

\subsection{Configuration Insight: Qwen3-4B Batch Size}
\label{sec:qwen-prompt-batch}

The strong result of \textsf{Qwen3-4B} relative to the larger \textsf{Qwen3-8B} and \textsf{Qwen3-14B} variants should be interpreted as a configuration-level finding. The 4B run used a more recent instruction-tuned checkpoint and a larger batch size of 32, whereas the 8B and 14B variants were run with batch sizes of 16 and 8, respectively.

To assess whether this configuration contributed to the result, two faster alternatives were evaluated: the same structured prompt with batch size 4, and a simple single-segment prompt that requested only one 0--100 score. The simple prompt is provided in \Cref{app:qwen-simple-prompt}. As \Cref{tab:qwen_batch} shows, both alternatives substantially increased throughput, but they also lost the ranking behavior that made \texttt{qwen3-4b} competitive.

\begin{table}[t]
  \centering
  \small
  \setlength{\tabcolsep}{5pt}
  \renewcommand{\arraystretch}{1.08}
  \begin{tabular}{lrrrr}
    \toprule
    Configuration & Dir./hour $\uparrow$ & Wins $\uparrow$ & Avg. rank $\downarrow$ & Mean score $\uparrow$ \\
    \midrule
    Batch 32 & $\sim$20 & 11,559 & 2.44 & 0.6160 \\
    Batch 4 & $\sim$60 & 4 & 7.24 & 0.4281 \\
    Simple & $\sim$160 & 0 & 8.63 & 0.3897 \\
    \bottomrule
  \end{tabular}
  \caption{Prompt and batch-size sensitivity for Qwen3-4B. Dir./hour denotes FLORES-200 direction-equivalents processed per hour. Statistics come from an augmented comparison that adds the two faster Qwen3-4B settings to the original benchmark, and wins are computed within that augmented method pool.}
  \label{tab:qwen_batch}
\end{table}

One plausible interpretation is that the larger batch provided local calibration context.
When 32 source--translation pairs were presented together, the model could observe a wider range of examples before assigning scores, which may have encouraged more stable use of the 0--100 scale.
This interpretation is consistent with evidence that LLM outputs vary with prompt context and that contextual calibration can reduce some forms of variance \citep{zhao2021calibrate}; it is also compatible with batch-prompting work showing that batch size and item order can affect results \citep{cheng2023batch_prompting,lin2024batchprompt}.
For this benchmark, batching is therefore not only an efficiency parameter but also part of the calibration behavior of the evaluator.

\end{document}